%% file: arxiv.tex
\documentclass{article}
\usepackage{amd}
\input{math_commands.tex}

\newcommand{\AMDlogo}{\includegraphics[height=0.28in]{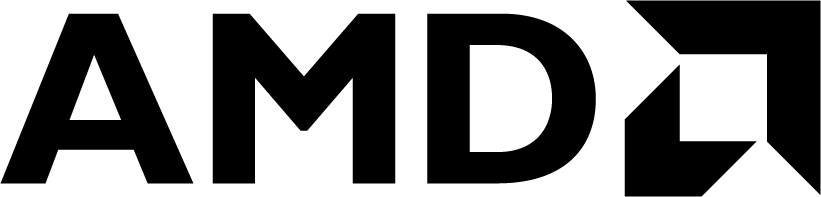}}
\newcommand{\CoralEmoji}{\raisebox{-0.14em}{\includegraphics[height=0.94em]{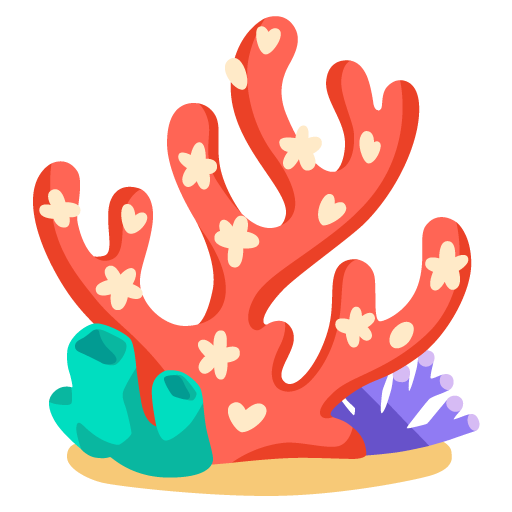}}}

\usepackage{booktabs}
\usepackage{graphicx}
\usepackage{svg}
\usepackage{algorithm}
\usepackage{algorithmic}
\usepackage{hyperref}
\usepackage{url}
\usepackage{placeins}
\usepackage{xcolor}
\usepackage{colortbl}
\usepackage{tabularx}

\definecolor{coral}{HTML}{FF8559}
\definecolor{reusablehighlight}{RGB}{255,241,235}
\definecolor{adaptedhighlight}{RGB}{255,230,220}

\begin{document}
\onecolumn
% Short running title:
\amdtitlerunning{CORAL-KV: Compact Cross-Layer Prediction and Residual Coding
for Autoregressive Video KV Caches}

\amdtitle{\CoralEmoji\ CORAL-KV: Compact Cross-Layer Prediction and Residual Coding
for Autoregressive Video KV Caches}

\icmlsetsymbol{equal}{*}

  \begin{amdauthorlist}
    \amdauthor{Albert Ge}{amd,uni}
    \amdauthor{Akash Haridas}{amd}
    \amdauthor{Utkarsh Saxena}{amd,cor} 
    \amdauthor{Ruijing Yang}{amd,cor} 
    \amdauthor{Ian Colbert}{amd}
    \amdauthor{Frederic Sala}{uni}
    \amdauthor{Vikram Appia}{amd}
    \amdauthor{Emad Barsoum}{amd}
  \end{amdauthorlist}
  
  \amdaffiliation{amd}{Advanced Micro Devices Inc. (AMD)}
  \amdaffiliation{uni}{University of Wisconsin-Madison}

  \amdcorrespondingauthor{Utkarsh Saxena}{utkarsh.saxena@amd.com}
  \amdcorrespondingauthor{Ruijing Yang}{ruijing.yang@amd.com}

  \amdkeywords{Video Generation, KV cache quantization, cross-layer prediction}
  \vskip 0.3in

  % is text to display at the start of the footnote. The \amdEqualContribution

% Print affiliations inline (no footnote block):
\amdPrintAffiliationsInline
% Public arXiv version displays authors and affiliations.

% Uncomment only for the camera-ready version.
% \iclrfinalcopy

\newcommand{\method}{\textsc{Coral-KV}}

\begin{abstract}
Autoregressive video diffusion models maintain a growing attention key--value
(KV) cache, making memory a key bottleneck for long video generation.
Recently, clustering-based quantization has emerged as an effective method to
compress video KV caches by grouping similar key and value vectors then
quantizing residual errors. However, when
later frames attend to the compressed history, small reconstruction errors can
compound as the video grows. Beyond the within-layer similarity exploited by
clustering, we find strong dependencies across cache layers in video models.
This motivates
\method{}---\textbf{C}alibrated
\textbf{O}ffline, \textbf{R}esidual-coded, and \textbf{A}dapted
\textbf{L}ocally---a KV-cache compression framework that uses cache states in
each layer to predict keys and values in subsequent layers, then quantizes the
remaining prediction residuals.
We construct compact block-diagonal-plus-low-rank (block-DPLR) predictors:
block-diagonal components model channel relationships within each
attention head, while a global low-rank term models dependencies between heads.
\method{} exposes a flexible fidelity--memory trade-off: it can reuse its
predictors across videos or refit lightweight block-diagonal predictors to the
current video for higher fidelity. Across extensive evaluations on both
HY-WorldPlay and MAGI-1, our reusable configuration
improves PSNR over QVG-4 by 0.32~dB and 4.95~dB,
respectively, while using 48\% and 37\% less cache memory.
Against QVG-Pro, our video-adapted configuration
improves MAGI-1 PSNR by 3.77~dB while using 33.7\% less cache memory.
\method{} also maintains higher reconstruction fidelity over extended
generation and reduces temporal discontinuities while retaining comparable
generation latency.
\end{abstract}

\section{Introduction}

Recent advances in video generation point toward a powerful new paradigm:
learned simulators that can model and seemingly reason about the dynamics of the physical
world
\citep{deepmind2025veo3,kong2024hunyuanvideo,polyak2024moviegen,wan2025,
wiedemer2025videomodels}. Beyond content creation, such systems could support
interactive environments, embodied agents, and general-purpose visual reasoning
\citep{he2025matrixgame2,xiao2025worldmem}. Autoregressive video models are
particularly well suited to realizing this promise: by generating video
incrementally while conditioning each new segment on the preceding history,
they support causal interaction and long-horizon coherence
\citep{huang2025selfforcing,sandai2025magi1,yang2025longlive}.

Despite their promise, autoregressive video models face a severe memory
bottleneck: the attention key--value (KV) cache grows with video
duration, spatial resolution, and model
depth, and can quickly dominate GPU memory
\citep{li2026packcache,samuel2026fastar}. Consequently, memory constraints limit
both where these models can be used and how much history they can retain,
even though longer context is critical for preserving
identity, scene geometry, and motion over long rollouts
\citep{xiao2025worldmem,hong2025relic}.

\begin{figure}[!t]
    \centering
    \includegraphics[width=\textwidth]{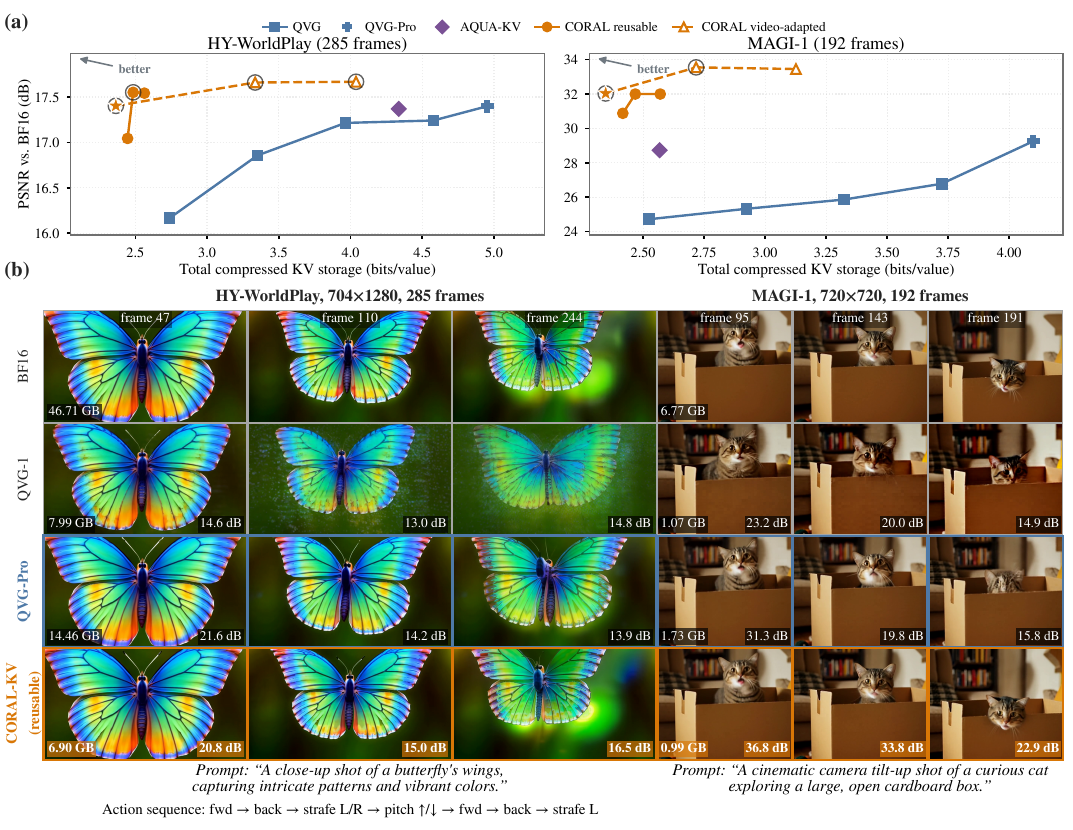}
    \vspace{-0.5em}
    \caption{\textbf{CORAL-KV improves the memory--quality frontier.}
    (a) average PSNR versus total compressed KV storage across 32 prompts per model,
    using the same storage accounting as in
    Table~\ref{tab:main-quality-memory}; stars mark our selected reusable codec.
    (b) Qualitative comparisons of BF16, QVG-1, QVG-Pro,
    and reusable CORAL-KV on matched frames.}
    \label{fig:teaser}
    \vspace{-1.5em}
\end{figure}

Prior work has shown that video KV caches can be compressed by exploiting
spatiotemporal redundancy among video tokens
\citep{xi2025svg,yang2025svg2}. In particular, Quant VideoGen (QVG) realizes
this idea by separately clustering similar key vectors and similar value
vectors within each layer, representing each cluster with a centroid, and
quantizing the residual errors to low-bit precision~\citep{xi2026qvg}. Over long generation
horizons, however, we observe that small quantization errors early in generation can
propagate through later frames, producing substantial reconstruction drift
(Figure~\ref{fig:teaser}).

Instead of clustering similar vectors within each cache layer, we find and exploit strong
cross-layer dependencies in video KV caches. These dependencies hold across
two distinct video architectures (Figure~\ref{fig:motivation}). AQUA-KV observed similar
dependencies in large language model (LLM) KV caches and used dense linear predictors with residual
quantization~\citep{shutova2025aquakv}. Directly transferring this approach to
video models, however, introduces new bottlenecks: dense predictor memory
scales quadratically with the KV hidden dimension, and applying these
predictors at every layer adds substantial inference overhead.
We address these bottlenecks with \method{}, a practical KV-cache compression
framework that combines compact block-diagonal-plus-low-rank (block-DPLR) predictors with
low-bit residual quantization.
In summary, our contributions are as follows:
\begingroup
\setlength{\leftmargini}{1.5em}
\begin{itemize}
    \setlength{\topsep}{2pt}
    \setlength{\itemsep}{0pt}
    \setlength{\parsep}{0pt}
    \setlength{\parskip}{0pt}
    \item We introduce \method{}, which uses
    %, a KV-cache compression framework that makes cross-layer prediction practical with
    block-DPLR predictors which require
    % that preserve within- and cross-head information using
    $13.1\times$ and $3.0\times$ \textit{fewer} matrix parameters than dense prediction on HY-WorldPlay and MAGI-1, respectively. The block-DPLR predictors can be calibrated once offline for
    reuse across videos or adapted online by refitting only block-diagonal
    components per cache block for higher fidelity.
    \item We co-design prediction and residual quantization by calibrating
    predictors on reconstructed cache states and tailoring blockwise Hadamard
    transforms and 2-bit vector quantization to their residuals. Our direct
    factorized reconstruction avoids dense predictor matrices, reducing memory
    and inference overhead.
    \item Through extensive evaluations on two video generators, we show that \method{} improves
    visual fidelity and temporal stability over matched baselines; the
    reusable configuration has comparable generation latency.
\end{itemize}
\endgroup

\section{Related work}
\label{sec:related-work}

We focus here on the three areas most directly related to \method{}: video-cache
compression and attention reduction, LLM KV compression, and low-bit
quantization.
Appendix~\ref{sec:appendix-related-work} covers broader related work.

\textbf{Video-cache compression and attention reduction.}
We directly compare \method{} to QVG, which fits per-head centroid codebooks
online for each cache block and progressively quantizes deviations from them
\citep{xi2026qvg}. Contemporaneous QuantWM uses sensitivity-aware clustering
and low-rank attention compensation to reduce quantization artifacts in video
world models~\citep{zhao2026quantwm}. Other video methods prune or compact
tokens, sparsify
attention, replace dense attention state with low-rank representations, or
absorb history into online parameters~\citep{li2026packcache,xi2025svg,yang2025svg2,fu2026ispa,ji2026forcingkv,yesiltepe2026videomla,zhao2026densitykv}. These approaches change which historical tokens remain accessible
or how attention uses them; \method{} retains every position while reducing
the memory needed to store its KV values.

\textbf{LLM KV compression and cross-layer prediction.}
Prior LLM methods quantize or correct cached state to reduce its memory
footprint~\citep{he2024zipcache,hooper2024kvquant,kang2024gear,liu2024kivi}.
We find cross-layer predictability in video caches and show that predictors
calibrated offline remain effective across videos. We exploit these findings
with compact block-DPLR predictors and low-bit residual coding. AQUA-KV reports
similar cross-layer predictability in LLM caches and uses dense predictors
with residual quantization~\citep{shutova2025aquakv}. Dense maps scale
quadratically with channel width, making them costly for wide video caches;
our factorized predictors reduce memory and inference cost in comparison.

\textbf{Quantization and residual coding.}
Prior quantization methods use rotations to distribute outliers and
improve low-bit coding~\citep{ashkboos2024quarot,liu2025spinquant}. We apply
fixed Hadamard transforms to cross-layer prediction residuals and choose how
many residual channels each transform mixes using our quantization and
video-quality ablations (Figure~\ref{fig:quantizer-selection};
Appendix~\ref{sec:appendix-residual-quantization}). We then encode the
transformed residuals with the fixed, Gaussian-tuned 2-bit EDEN-2 vector
codebook~\citep{vargaftik2022eden,malinovskii2025higgs}.

\section{Problem setting and motivation}

Video KV caches are extremely memory-intensive. In an autoregressive video
model with $L$ attention layers, KV channel width $C$, and $T$ retained patch
tokens, storing both keys and values in BF16 requires $2LTC\times16$ bits. For
a 285-frame, near-720p HY-WorldPlay video, retaining every completed BF16
cache block would require 46.71~GB (Table~\ref{tab:main-quality-memory}).
This highlights the critical need to compress cache memory
to enable long-horizon video generation.

\noindent\textbf{Quantizing cache blocks can compound reconstruction errors.}
One way to reduce video KV-cache memory is to quantize previously generated
keys and values. During generation, KV states for new visual tokens accumulate
in a \emph{cache block}. When the block reaches its token limit, it is compressed
and retained as history while generation continues with a new block. QVG fits
centroid codebooks \emph{online} for each filled block, clustering similar
keys and values within each layer before quantizing their deviations from the
centroids~\citep{xi2026qvg}.

\begin{figure*}[t]
    \centering
    \includegraphics[width=\textwidth]{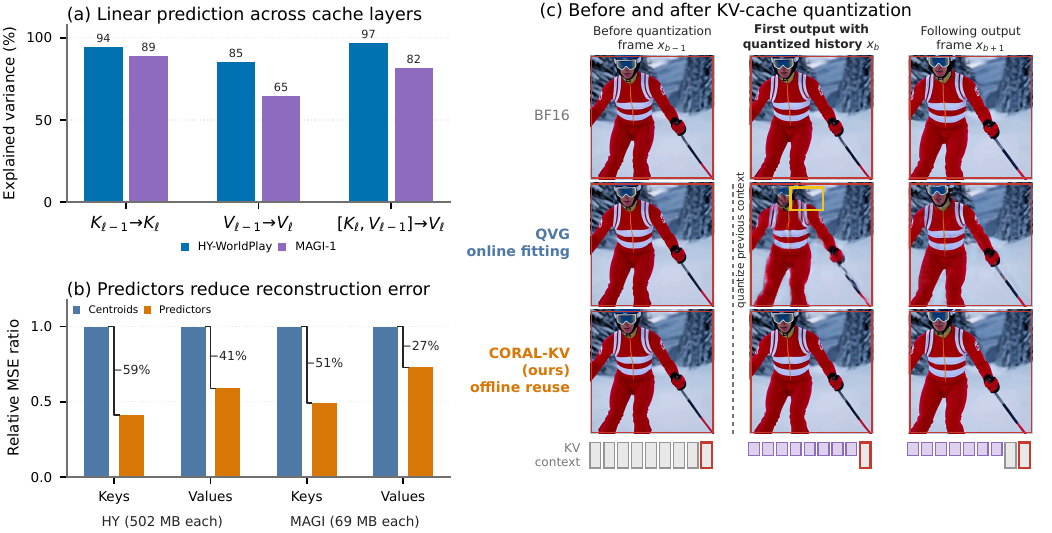}
    \caption{\textbf{Cross-layer prediction and quantization-boundary effects.}
    (a) Linear maps predict current-layer keys from previous-layer keys and
    current-layer values from previous-layer values. Adding current-layer keys
    improves value prediction. The key and augmented value maps explain over
    80\% of the variance on both models.
    (b) We compare reconstruction MSE from predictors and centroids on the same
    keys and values before residual quantization. Bars show relative MSE normalized
    to the centroid result. Predictors significantly reduce reconstruction
    MSE compared with centroids. (c) QVG fits video-specific
    centroids online when a cache block reaches its token limit, whereas CORAL-KV reuses
    offline-calibrated predictors. The yellow box highlights facial distortion
    in QVG's first frame generated from compressed history; visible degradation
    also appears in the clothing and ski pole.}
    \label{fig:motivation}
\end{figure*}

Although this can substantially reduce cache size, we observe that it
introduces small reconstruction errors early in the generated video, which can
become magnified as the video grows longer, manifesting in our qualitative
examples as oversmoothing, blended structures, and geometric drift
(Figure~\ref{fig:teaser}; also see Appendix
Figures~\ref{fig:additional-long-range}--\ref{fig:additional-long-range-pair-2}).
Once a filled cache block is quantized, the next generated frame is the first
to attend to the newly
compressed history. This transition can cause an abrupt quality drop and a
visible temporal discontinuity (Figure~\ref{fig:motivation}(c)). We call this
effect quantization-boundary discontinuity (QBD) and define its measurement in
Section~\ref{sec:experimental-methodology}.

\noindent\textbf{Cross-layer structure and reusable predictors.}
We first examine dependencies across layers. In HY-WorldPlay and MAGI-1, we train linear
maps to predict a layer's keys and values from preceding-layer cache states
(Figure~\ref{fig:motivation}(a)). Predictors fitted on one set of videos
remain effective on held-out videos, enabling \emph{offline} calibration and
reuse across videos.
Let $K_\ell$ and $V_\ell$ denote the unquantized keys and values at layer
$\ell$. We compare three tokenwise linear-map structures: keys to keys
($K_{\ell-1}\!\to K_\ell$), values to values
($V_{\ell-1}\!\to V_\ell$), and keys and values to values
($[K_\ell,V_{\ell-1}]\!\to V_\ell$).
For the key-to-key map, we report 94\% of explained variance on HY-WorldPlay and
89\% on MAGI-1, compared to only about 80\% explained variance on
Llama-3.2-3B~\citep{shutova2025aquakv}.  
Adding current-layer keys raises explained variance for value prediction from
85\% to 97\% on HY-WorldPlay and from 65\% to 82\% on MAGI-1
(Figure~\ref{fig:motivation}(a)). We therefore reconstruct keys first, then
use the reconstructed keys with preceding-layer values to predict current-layer
values. Because preceding-layer keys already predict current keys well, we omit
preceding-layer values from the key predictor to avoid doubling its input width
and reduce predictor memory.

We next compare our predictors (further discussed in Section~\ref{sec:coral-kv}) with centroid
codebooks at matched parameter storage (Figure~\ref{fig:motivation}(b)). To
match the storage of predictor weights and centroid vectors, we use centroid
codebooks with 705 centroids for each key or value head at each layer on
HY-WorldPlay and 513 on MAGI-1. In comparison, predictors reduce relative MSE by 59\% for
keys and 41\% for values on HY-WorldPlay, and by 51\% and 27\% on MAGI-1.
The full protocol is in Appendix~\ref{sec:appendix-equal-parameter-storage}.
While we do not offer a formal explanation for this highly accurate linear predictability, a plausible
intuition is that adjacent layers process the same token history and retain
shared structure; the held-out curves and cross-video reuse are consistent
with this hypothesis but do not establish it
(Section~\ref{sec:limitations}).

\section{CORAL-KV}
\label{sec:coral-kv}

\subsection{Reconstruction overview}
\label{sec:coral-overview}

\begin{figure*}[!t]
  \centering
  \includegraphics[width=0.96\textwidth]{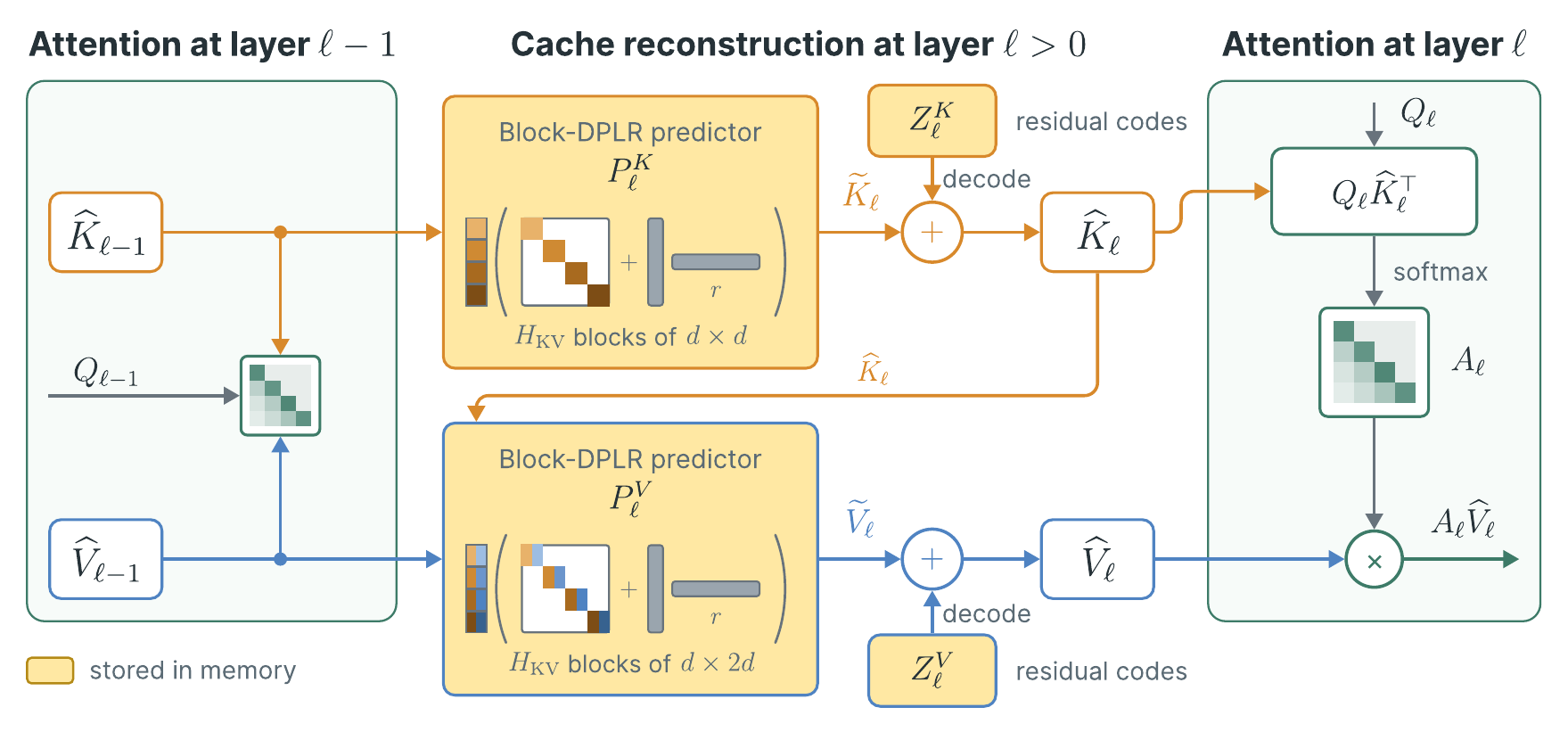}
  \vspace{-0.6em}
  \caption{\textbf{CORAL-KV reconstruction.} At layer $\ell>0$, reconstructed
  cache states from layer $\ell-1$ feed the predictors. The key predictor maps
  $\widehat K_{\ell-1}$ to $\widetilde K_\ell$; adding the quantized key
  residual gives $\widehat K_\ell$. The value predictor maps
  $[\widehat K_\ell,\widehat V_{\ell-1}]$ to $\widetilde V_\ell$; adding the
  quantized value residual gives $\widehat V_\ell$. Both predictors use a
  block-diagonal component and a rank-$r$ low-rank component.}
  \label{fig:coral-system}
  \vspace{0.3em}
  \begin{minipage}{\textwidth}
\begin{algorithm}[H]
  \caption{Block-DPLR calibration per layer}
  \label{alg:block-dplr-calibration}
  \small
  \begin{algorithmic}[1]
    \REQUIRE Inputs $X^K_\ell=\widehat K_{\ell-1}$ and
        $X^V_\ell=[\widehat K_\ell,\widehat V_{\ell-1}]$;
        targets $K_\ell,V_\ell$; rank $r$; three fitting rounds
    \ENSURE $P^S_\ell(x)=D^S_\ell x+A^S_\ell(B^S_\ell x)$ for
        $S\in\{K,V\}$
    \STATE Initialize $(A^K_\ell B^K_\ell)^{(0)}=0$ and
        $(A^V_\ell B^V_\ell)^{(0)}=0$
    \FOR{$t=1,2,3$}
        \STATE \textit{// Fit the block-diagonal maps to the remaining error}
        \STATE $R_D^K\gets K_\ell-(A^K_\ell B^K_\ell)^{(t-1)}X^K_\ell$;
            $R_D^V\gets V_\ell-(A^V_\ell B^V_\ell)^{(t-1)}X^V_\ell$
        \STATE $D_\ell^{K,(t)}\gets\arg\min_{D\in\mathcal D^K}
            \lVert DX^K_\ell-R_D^K\rVert_F^2$;
            $D_\ell^{V,(t)}\gets\arg\min_{D\in\mathcal D^V}
            \lVert DX^V_\ell-R_D^V\rVert_F^2$
        \STATE \textit{// Fit the low-rank maps to the remaining error}
        \STATE $R_{AB}^K\gets K_\ell-D_\ell^{K,(t)}X^K_\ell$;
            $R_{AB}^V\gets V_\ell-D_\ell^{V,(t)}X^V_\ell$
        \STATE $(A_\ell^{S,(t)},B_\ell^{S,(t)})\gets
            \arg\min_{\operatorname{rank}(AB)\le r}
            \lVert ABX^S_\ell-R_{AB}^S\rVert_F^2$ for $S\in\{K,V\}$
    \ENDFOR
    \RETURN $P_\ell^K,P_\ell^V$
  \end{algorithmic}
\end{algorithm}
  \end{minipage}
  \vspace{-0.8em}
\end{figure*}

Figure~\ref{fig:coral-system} illustrates how \method{} reconstructs each layer
of a compressed cache block. Once a block is full, we encode its KV states and
retain quantized codes and scales. When later chunks attend to that block, we
reconstruct its keys and values layer by layer. At layer $\ell>0$, a tokenwise
linear map $P^K_\ell$ predicts keys from reconstructed keys at layer $\ell-1$.
A second map $P^V_\ell$ predicts values from reconstructed keys at layer $\ell$
and reconstructed values at layer $\ell-1$.
Section~\ref{sec:coral-design} gives their compact parameterization. We augment
both predictions with quantized residuals, where $\mathcal Q_{\mathrm R}$ denotes
the residual quantizer formally defined in Appendix~\ref{sec:appendix-quantizer-specification}:
\begin{align}
\text{\scriptsize predicted keys: }\widetilde K_\ell&=P^K_\ell(\widehat K_{\ell-1}),&
\text{\scriptsize reconstructed keys: }\widehat K_\ell&=\widetilde K_\ell+
  \mathcal Q_{\mathrm R}(K_\ell-\widetilde K_\ell),\nonumber\\
\text{\scriptsize predicted values: }\widetilde V_\ell&=P^V_\ell([\widehat K_\ell,\widehat V_{\ell-1}]),&
\text{\scriptsize reconstructed values: }\widehat V_\ell&=\widetilde V_\ell+
  \mathcal Q_{\mathrm R}(V_\ell-\widetilde V_\ell).
\label{eq:coral-predictive-reconstruction}
\end{align}
Here, brackets concatenate channels. The predictors do not receive original keys
or values as inputs; they operate only on reconstructed tensors available to
the decoder. Because layer zero has no preceding layer, we quantize its keys
and values directly. In our ablations, we find that INT4 is sufficient
to preserve video fidelity for layer zero on both models
(Appendix~\ref{sec:appendix-configuration-selection};
Figure~\ref{fig:anchor-selection}).
Appendix~\ref{sec:appendix-quantizer-specification} gives the full
encoder--decoder specification.

\FloatBarrier
\subsection{Compact predictors and residual quantization}
\label{sec:coral-design}

\begin{figure*}[t]
    \centering
    \includegraphics[width=\textwidth]{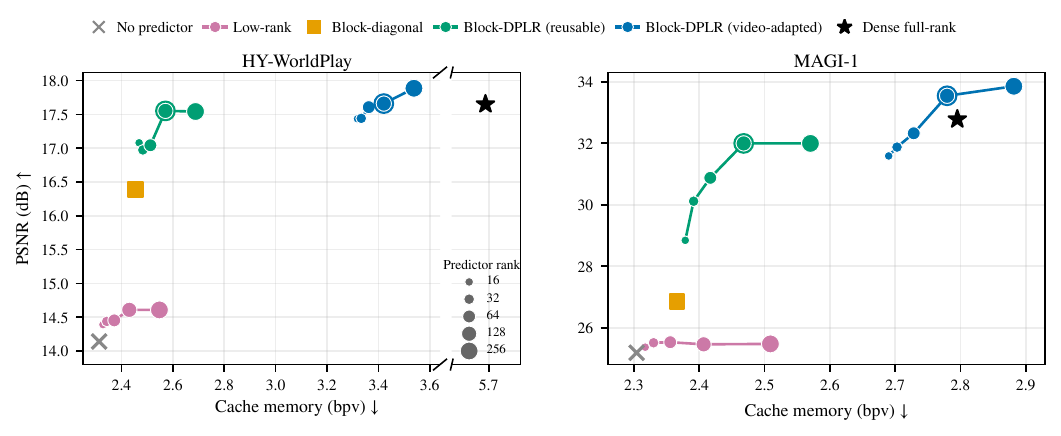}
    \caption{\textbf{Predictor-structure ablation.} Block-DPLR outperforms
    block-diagonal and parameter-matched low-rank predictors. Reusable
    block-DPLR shares offline predictors across videos. Video-adapted
    block-DPLR keeps its low-rank factors fixed and refits block-diagonal terms
    for each filled cache block, improving fidelity at higher memory cost.
    The dense control is also fitted offline.}
    \label{fig:predictor-selection}
\end{figure*}

Directly using dense linear maps for cross-layer prediction can erase the
memory savings from cache compression. For KV channel width $C$, a dense key predictor
contains $C^2$ parameters and a dense value predictor contains $2C^2$.
For~instance,~HY-WorldPlay has $C=6144$,
so storing these predictors in BF16 requires 6.57~GB across its predicted
layers. In our
evaluated setting, the predictors alone require nearly twice as much memory as the
compressed cache, while applying them at every layer increases generation time
by 71.7\% relative to BF16 (Table~\ref{tab:main-quality-memory}). A practical KV compressor therefore
requires a predictor that retains expressiveness without quadratic
memory and computation.

We propose a block-diagonal-plus-low-rank (block-DPLR) predictor with a
block-diagonal component $D_\ell$ and a rank-$r$ component $A_\ell B_\ell$.
Let $H_{\mathrm{KV}}$ denote the number of KV attention heads and
$d=C/H_{\mathrm{KV}}$ their width. For keys, each head-aligned block
$D_{\ell,h}^K\in\mathbb R^{d\times d}$ maps the corresponding preceding-layer
key head, while $A_\ell^K\in\mathbb R^{C\times r}$ and
$B_\ell^K\in\mathbb R^{r\times C}$ incorporate information across heads. For
values, each $D_{\ell,h}^V\in\mathbb R^{d\times 2d}$ maps the corresponding
current-layer key and preceding-layer value heads, while
$A_\ell^V\in\mathbb R^{C\times r}$ and $B_\ell^V\in\mathbb R^{r\times 2C}$
provide the cross-head component.
Because $D_{\ell}$ is block diagonal over attention heads, fitting it reduces to one least-squares solve per head, which can be parallelized. We revisit this property in the predictor ablations below, and in Appendix~\ref{sec:appendix-efficient-inference}.

Our proposed predictor structure sums the block-diagonal and low-rank components, e.g. $P_{\ell}(x) = D_{\ell}(x) + A_{\ell}B_{\ell}(x)$.  We fit this sum by alternating block minimization~\citep{tseng2001block}
(Algorithm~\ref{alg:block-dplr-calibration}). Each update fits one component to the
target after subtracting the other component's prediction, so
the two components model complementary information. We write $\mathcal D^K$
and $\mathcal D^V$ for the corresponding head-aligned block-diagonal matrix
families.

The key and value predictors use $H_{\mathrm{KV}}d^2+2Cr$ and
$2H_{\mathrm{KV}}d^2+3Cr$ matrix parameters, respectively. At rank $r=128$,
block-DPLR uses $13.1\times$ fewer matrix parameters than dense prediction on
HY-WorldPlay and $3.0\times$ fewer on MAGI-1.
At inference, we apply $D_\ell x$ in head batches and compute the low-rank
term $A_\ell(B_\ell x)$ without forming a dense predictor. We develop fused GPU kernels that
decode residuals, apply these factors, and write reconstructed states directly
into the attention layout. Appendix~\ref{sec:appendix-efficient-inference}
provides implementation details.

\noindent\textbf{Predictor structure.}
The alternating fit calibrates one key and one value predictor per layer from
cache tokens pooled across 32 diverse MovieGen Extended
videos~\citep{polyak2024moviegen}; \emph{reusable} \method{} shares these
offline-calibrated predictors across videos.
Appendix~\ref{sec:appendix-calibration-scaling} varies the number of
calibration prompts while keeping the predictor structure fixed.
\emph{Video-adapted} \method{} keeps the offline low-rank factors but refits
block-diagonal terms online for each filled cache block of the current video.
We compare offline block-diagonal,
low-rank, block-DPLR, and dense predictors across parameter budgets, alongside
a no-predictor control. Figure~\ref{fig:predictor-selection} shows that
block-DPLR outperforms block-diagonal and parameter-matched low-rank predictors
on both models. The offline dense predictor gains only 0.10 and 0.79~dB over
reusable rank-128 block-DPLR on HY-WorldPlay and MAGI-1, respectively, while
requiring substantially more predictor memory.

This per-block refitting improves fidelity at a memory cost. Online low-rank refitting gives
similar fidelity, but its global update takes $2.6\times$ as long to fit as the
parallel headwise block-diagonal
update. We therefore keep the offline low-rank factors fixed in the
video-adapted configuration, which gives the highest quality in
Figure~\ref{fig:predictor-selection}. Appendix~\ref{sec:appendix-configuration-selection}
reports the full ablations and confidence intervals.

\begin{figure*}[t]
    \centering
    \includegraphics[width=0.98\textwidth]{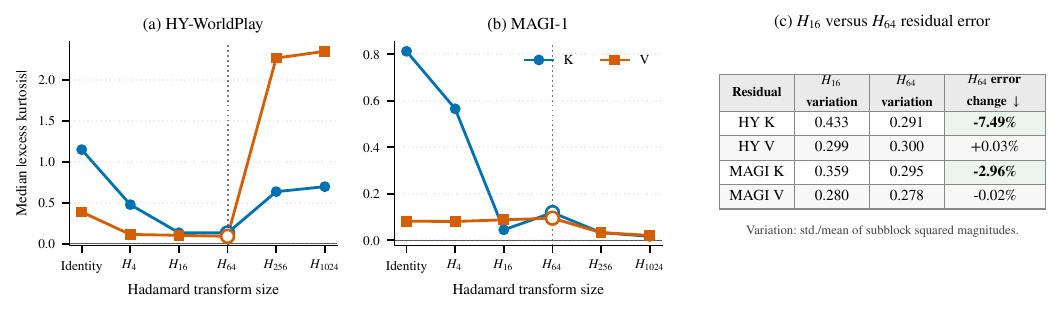}
    \caption{(a--b) Median absolute excess kurtosis across layers and two prompts; lower
    is closer to a unit Gaussian. $H_n$ transforms each non-overlapping group of
    $n$ residual channels independently. Wider transforms have model-dependent
    effects beyond $H_{64}$. (c) Comparison at one shared scale per 64 values.
    ``Variation'' is the standard deviation divided by the mean of the squared
    magnitudes of the four 16-value subblocks; lower values indicate more
    uniform magnitudes.}
    \label{fig:quantizer-selection}
\end{figure*}

\noindent\textbf{Quantizing residuals.}
To preserve the predictor's reconstruction gains within a two-bit residual
budget, we tailor the quantizer to the resulting prediction errors. At this low
precision, concentrated coordinates make quantization sensitive to
outliers~\citep{sanjeet2026mixquant}. Hadamard rotations can spread outliers
across channels~\citep{ashkboos2024quarot}; transformed LLM weights and
activations can also become approximately
Gaussian~\citep{malinovskii2025higgs,egiazarian2026mrgptq}.
The effect of a fixed block Hadamard transform depends on input geometry and
transform width~\citep[Proposition~3.2 and Corollary~3.3]{sanjeet2026mixquant}. On our prediction
residuals, wider transforms can instead concentrate correlated structure in a
few coefficients.
It is therefore unclear how many channels should be transformed together.
We test independent transforms over groups of $n=4$ to 1024 channels, denoted
$H_n$. Figure~\ref{fig:quantizer-selection}(a--b) measures absolute excess
kurtosis (zero for a Gaussian). $H_{16}$ and $H_{64}$ perform most consistently
across both models; larger groups help MAGI-1 but give HY-WorldPlay heavier
tails, especially for values.

Figure~\ref{fig:quantizer-selection}(c) compares $H_{16}$ and $H_{64}$ with one
scale per 64 residual values. $H_{16}$ transforms four 16-value groups;
$H_{64}$ transforms the entire group. $H_{64}$ makes key magnitudes more even
across the four groups, reducing key-residual quantization
error by 7.49\% on HY-WorldPlay and 2.96\% on MAGI-1; value-residual error
changes by less than 0.03\%. At the same bit allocation, $H_{64}$ improves mean
video PSNR by 0.69 and 0.25~dB, respectively. We use $H_{64}$ and assign each
normalized, transformed residual pair to one of 16 codewords in the fixed,
Gaussian-tuned EDEN-2 codebook~\citep{vargaftik2022eden}. An inverse transform reconstructs the
residuals before they are added in Eq.~\ref{eq:coral-predictive-reconstruction}.

To compare with QVG while assigning the same number of bits to residual codes
and scales, we target 2.125 bits per value. Each prediction residual uses a
two-bit code, leaving 0.125 bits per value for scales. For keys, we store one FP8 E4M3 scale per 64 residual values; for values, we store one FP16 scale per 128 residual values.
Our ablations show that keys are more sensitive than
values to sharing a scale across adjacent blocks, which motivates this
allocation. We specify the quantizer in
Appendix~\ref{sec:appendix-quantizer-specification} and report the full
ablations in Appendix~\ref{sec:appendix-residual-quantization}.

\section{Experimental methodology}
\label{sec:experimental-methodology}

\noindent\textbf{Models, data, and baselines.}
We evaluate \method{} on two autoregressive video generators. HY-WorldPlay
generates 285 frames at $704\times1280$ resolution using four denoising steps.
Due to resource constraints, its fidelity evaluations retain the most recent
48 latent frames of context. MAGI-1 generates 192 frames at $720\times720$ resolution using 16
denoising steps. We evaluate on 355 held-out MovieGen Extended
prompts~\citep{polyak2024moviegen} that are disjoint from those used for
calibration and model selection.

Baselines are BF16, QVG-1, QVG-4, QVG-Pro, and an AQUA-KV-based dense
cross-layer predictor. QVG-1 uses one centroid-fitting stage per cache block
and quantizes its residual in 64-value groups; QVG-4 adds three more centroid stages
with the same residual group size. QVG-Pro uses four stages and 16-value
residual groups~\citep{xi2026qvg}. It is our strongest quality baseline;
QVG-4 matches reusable \method{}'s residual bit allocation and serves as our
latency baseline. We evaluate reusable \method{} with offline-calibrated
predictors and video-adapted \method{} with block-diagonal components refit
per cache block.
Appendix~\ref{sec:appendix-evaluation-configurations} gives model and split details.

\begin{table*}[!t]
    \centering
    \caption{\textbf{\method{} improves fidelity with less total cache memory
    in matched comparisons against existing baselines on both models.}
    HY-WorldPlay memory counts all completed cache blocks in the 285-frame
    video; its fidelity evaluations retain the most recent 48 latent frames.
    Shared predictors are counted once at 16 videos.
    Latency is generation-time overhead relative to BF16; its measurement
    protocol appears in Appendix~\ref{sec:appendix-latency}.}
    \label{tab:main-quality-memory}
    \scriptsize
    \setlength{\tabcolsep}{0.4pt}
    \renewcommand{\arraystretch}{0.94}
    \begin{tabularx}{\textwidth}{@{}l*{6}{>{\centering\arraybackslash}X}@{}}
        \toprule
        & \multicolumn{3}{c}{Fidelity}
        & \multicolumn{2}{c}{Total memory (GB) $\downarrow$}
        & \multicolumn{1}{c}{Latency} \\
        \cmidrule(lr){2-4}\cmidrule(lr){5-6}\cmidrule(l){7-7}
        Method
        & PSNR $\uparrow$
        & SSIM $\uparrow$
        & LPIPS $\downarrow$
        & 1 video
        & 16 videos
        & Overhead $\downarrow$ \\
        \midrule
        \multicolumn{7}{l}{\textbf{HY-WorldPlay (285 frames)}} \\
        BF16
        & -- & -- & -- & 46.71 & 747.42 & $0.0\%$ \\
        QVG-1
        & 15.70 & 0.496 & 0.513
        & 7.99 & 127.91 & $+7.8\%$ \\
        QVG-4
        & 16.66 & 0.540 & 0.438
        & 13.36 & 213.82 & $+15.3\%$ \\
        QVG-Pro
        & 17.16 & 0.559 & 0.408
        & 14.46 & 231.34 & $+17.7\%$ \\
        AQUA-KV
        & 17.12 & 0.560 & 0.399
        & 12.66 & 104.03 & $+71.7\%$ \\
        \rowcolor{reusablehighlight}
        \method{} (reusable)
        & 16.98 & 0.551 & 0.406
        & \textbf{6.90} & \textbf{102.90} & $+8.7\%$ \\
        \rowcolor{adaptedhighlight}
        \method{} (video-adapted)
        & \textbf{17.38} & \textbf{0.568} & \textbf{0.392}
        & 9.74 & 148.26 & $+29.1\%$ \\
        \midrule
        \multicolumn{7}{l}{\textbf{MAGI-1 (192 frames)}} \\
        BF16
        & -- & -- & -- & 6.77 & 108.29 & $0.0\%$ \\
        QVG-1
        & 24.13 & 0.785 & 0.250
        & 1.07 & 17.09 & $+2.6\%$ \\
        QVG-4
        & 26.14 & 0.830 & 0.207
        & 1.58 & 25.20 & $+10.3\%$ \\
        QVG-Pro
        & 28.70 & 0.872 & 0.162
        & 1.73 & 27.74 & $+9.6\%$ \\
        AQUA-KV
        & 28.13 & 0.861 & 0.159
        & 1.09 & \textbf{14.26} & $+6.3\%$ \\
        \rowcolor{reusablehighlight}
        \method{} (reusable)
        & 31.09 & 0.899 & 0.127
        & \textbf{0.99} & 14.84 & $+6.8\%$ \\
        \rowcolor{adaptedhighlight}
        \method{} (video-adapted)
        & \textbf{32.47} & \textbf{0.913} & \textbf{0.111}
        & 1.15 & 17.35 & $+10.0\%$ \\
        \bottomrule
    \end{tabularx}
\end{table*}

\noindent\textbf{Quality evaluation.}
We measure reconstruction fidelity against matched BF16 generations using RGB
PSNR, SSIM~\citep{wang2004ssim}, and VGG-LPIPS~\citep{zhang2018lpips}.
We compress each filled cache block as a unit: 7,040 visual tokens on
HY-WorldPlay and 12,150 on MAGI-1. To stress-test sustained generation with
compressed history, we score only frames after the first compressed block begins
affecting generation. The first affected frames are 29 and 24, respectively
(zero-indexed), so fidelity is evaluated on frames 30--284 and 25--191.
To quantitatively evaluate the compression-boundary discontinuities in
Figure~\ref{fig:motivation}(c), we introduce quantization-boundary
discontinuity (QBD).

QBD embeds each decoded frame of video $x$ with DINO
ViT-B/16~\citep{caron2021dino}.
Let $h_t$ be the normalized embedding of frame $t$ and
$s_t=h_t^\top h_{t-1}$ the cosine similarity between adjacent frames. We
compare the least similar pair among the transition into frame $u$ and the next
two transitions with the video's typical pair, represented by the median
$m$ of its adjacent-frame similarities. The difference in similarity is
$d(u)=m-\min\{s_u,s_{u+1},s_{u+2}\}$; a large $d(u)$ means that at least one
of these three frame pairs is much less similar than a typical pair in the
same video. Let $\mathcal B$ contain the first decoded frame affected by each
quantized cache block. Let $\mathcal I$ contain all other starting frames for
which the three transitions fit within the video, except those within three
frames of $\mathcal B$.
We then compare the average scores at these two sets of locations:
\begin{equation}
\operatorname{QBD}(x)=
\frac{|\mathcal B|^{-1}\sum_{b\in\mathcal B}d(b)}
     {|\mathcal I|^{-1}\sum_{u\in\mathcal I}d(u)}.
\label{eq:qbd}
\end{equation}
Higher QBD indicates a more pronounced change when newly compressed history
first affects generation.
Table~\ref{tab:main-video-diagnostics} divides each method's average per-video
QBD by the average for its paired BF16 generations, measured at the same
scheduled frame indices. Appendix~\ref{sec:appendix-qbd-definition}
checks other numbers of transitions and distances from the first affected
frames. We additionally
report six VBench dimensions~\citep{huang2024vbench}.

\noindent\textbf{Memory and latency evaluation.}
For HY-WorldPlay, we calculate the total stored-cache memory for all completed
blocks of the 285-frame generation from measured per-block sizes, although the
fidelity runs retain only the most recent 48 latent frames. For MAGI-1, we
calculate the total cache memory used within its context limit. These totals
describe stored cache representations, not peak GPU memory. They include residuals, scales,
assignments, anchors, predictors or codebooks, and
video-specific parameters.
We measure generation time on 16 matched prompts using identical attention
implementations and include any video-specific fitting time.
Appendices~\ref{sec:appendix-full-history-memory} and~\ref{sec:appendix-latency}
give the memory accounting and latency protocol, respectively.

\section{Main results}
\label{sec:main-results}

\noindent\textbf{\method{} achieves higher fidelity with less cache memory.}
Table~\ref{tab:main-quality-memory} compares reusable \method{} with QVG-1
and QVG-4, and video-adapted \method{} with QVG-Pro. Against
QVG-1, reusable \method{} improves PSNR by 1.28~dB on HY-WorldPlay and
6.96~dB on MAGI-1, with 13.7\% and 7.5\% less cache memory, respectively.
At the same residual-bit allocation as QVG-4, it improves PSNR by
0.32~dB and 4.95~dB while using 48.4\% and 37.3\% less cache memory,
respectively, and lowers generation overhead on both models.
Increasing QVG from one to four centroid stages raises both PSNR and cache
memory, but QVG-4 still trails reusable \method{} in PSNR while using more
memory on both models.

Video-adapted \method{} improves all three fidelity metrics over QVG-Pro while
using less cache memory on both models, including a 3.77~dB gain
with 33.7\% less memory on MAGI-1 and 32.7\% less memory on HY-WorldPlay.
Across the QVG-4 and QVG-Pro
comparisons, paired 95\% bootstrap intervals for PSNR,
SSIM, and LPIPS gains exclude zero on both models.
The video-adapted option pays for its higher fidelity with online block fitting;
a direct matched comparison finds higher generation overhead than reusable
\method{} on both models (Appendix~\ref{sec:appendix-latency}).
We also compare \method{} with DensityKV, a recent video-cache token-selection
method, and find that \method{} significantly improves PSNR by 3.27--5.11~dB
at matched compressed-history memory
(Appendix~\ref{sec:appendix-token-selection}).

At 16 videos, shared predictors give reusable \method{} lower total memory
than QVG-1 and QVG-4 on both models. HY-WorldPlay's full-history total is
6.90~GB versus 13.36~GB for QVG-4 (Table~\ref{tab:main-quality-memory}),
pointing toward longer-context video generation.
The AQUA-KV comparison illustrates the value of
compact prediction: on wide HY-WorldPlay caches, video-adapted \method{} uses
23.1\% less full-history memory and cuts generation overhead from 71.7\% to 29.1\%.
On MAGI-1, our AQUA-KV baseline uses the published 1,024-value Hadamard
groups~\citep{shutova2025aquakv} with one FP16 scale per group. Its residual
format costs 2.016 rather than \method{}'s 2.125 bits per value, making its
16-video total slightly smaller (14.26 versus 14.84~GB). Reusable \method{}
uses less memory for one video and improves PSNR by 2.96~dB.
Appendix~\ref{sec:appendix-concurrency} gives the memory-sharing details.

\noindent\textbf{Lower boundary discontinuity with similar VBench scores.}
Table~\ref{tab:main-video-diagnostics} shows that both configurations reduce
quantization-boundary discontinuity relative to QVG-Pro on both models. This
advantage persists across all 12 tested settings, which vary the number of
frame transitions considered and how far comparison windows must be from the
first affected frames (Appendix~\ref{sec:appendix-qbd-definition}).
Video-adapted \method{} nearly
halves QBD on MAGI-1 (1.328 versus 2.527). Both configurations
use 32.7--52.3\% less single-video cache memory
(Table~\ref{tab:main-quality-memory}) while achieving broadly similar VBench
scores. The largest observed VBench gain is HY-WorldPlay imaging quality, up
1.1--1.6 points. Appendix
Figures~\ref{fig:additional-boundaries}--\ref{fig:additional-boundaries-magi}
show further examples of the first compression boundary on both models.

Although QVG has higher VBench motion-smoothness and flicker scores on
HY-WorldPlay, unusually smooth frames can produce such scores. To check for
suppressed temporal variation, we measure the mean absolute pixel difference
between adjacent post-quantization frames on 32 prompts after
grayscale downsampling, normalized to each video's paired BF16 generation.
QVG shows 13\% less frame-to-frame change than BF16, whereas \method{} nearly
matches the BF16 frame-to-frame change. Thus QVG's higher scores partly
reflect suppressed variation (Appendix~\ref{sec:appendix-vbench-dynamics});
Appendix Figures~\ref{fig:additional-long-range}--\ref{fig:additional-long-range-pair-2}
show related examples of smoothing and detail loss.

\begin{table*}[!t]
    \centering
    \caption{\textbf{\method{} improves boundary stability with less cache
    memory and broadly similar VBench scores relative to QVG-Pro on both
    models.} Cache memory (GB) is the single-video total from
    Table~\ref{tab:main-quality-memory}.
    All results use 355 prompts.}
    \label{tab:main-video-diagnostics}
    \scriptsize
    \setlength{\tabcolsep}{1.5pt}
    \renewcommand{\arraystretch}{0.92}
    \begin{tabularx}{\textwidth}{@{}l*{8}{>{\centering\arraybackslash}X}@{}}
        \toprule
        Method
        & Cache $\downarrow$
        & QBD $\downarrow$
        & \multicolumn{6}{c}{VBench score $\uparrow$} \\
        \cmidrule(lr){4-9}
        &
        &
        & Subject
        & Background
        & Imaging
        & Aesthetic
        & Motion
        & Flicker \\
        \midrule
        \multicolumn{9}{l}{\textbf{HY-WorldPlay}} \\
        BF16
        & 46.71 & 1.000 & 89.153 & 92.821 & 69.434 & 61.860 & 98.208 & 96.073 \\
        QVG-1
        & 7.99 & 3.346 & 84.399 & 91.032 & 57.982 & 59.959
        & \textbf{98.508} & \textbf{97.263} \\
        QVG-4
        & 13.36 & 1.386 & 88.436 & 92.395 & 66.869 & 61.932 & 98.328 & 96.683 \\
        QVG-Pro
        & 14.46 & 1.158 & \textbf{89.317} & \textbf{92.932} & 68.040
        & \textbf{61.965} & 98.274 & 96.440 \\
        AQUA-KV
        & 12.66 & \textbf{1.003} & 88.748 & 92.557 & 69.513 & 61.737 & 98.128 & 96.010 \\
        \rowcolor{reusablehighlight}
        \method{} (reusable)
        & \textbf{6.90} & 1.022 & 88.555 & 92.558 & \textbf{69.604} & 61.925 & 98.105 & 95.998 \\
        \rowcolor{adaptedhighlight}
        \method{} (video-adapted)
        & 9.74 & 1.026 & 88.791 & 92.700 & 69.166 & 61.874 & 98.155 & 96.085 \\
        \midrule
        \multicolumn{9}{l}{\textbf{MAGI-1}} \\
        BF16
        & 6.77 & 1.000 & 97.218 & 97.302 & 64.205 & 64.904 & 99.433 & 98.976 \\
        QVG-1
        & 1.07 & 3.499 & 96.538 & 96.820 & 63.488 & 63.975 & 99.298 & 98.849 \\
        QVG-4
        & 1.58 & 3.194 & 96.828 & 97.012 & 63.637 & 64.445 & 99.361 & 98.893 \\
        QVG-Pro
        & 1.73 & 2.527 & 97.203 & 97.227 & 64.188 & 65.030 & 99.405 & 98.964 \\
        AQUA-KV
        & 1.09 & 3.537 & 97.092 & 97.205 & 64.085 & 64.778 & 99.411 & 98.950 \\
        \rowcolor{reusablehighlight}
        \method{} (reusable)
        & \textbf{0.99} & 2.150 & 97.329 & 97.297 & \textbf{64.647}
        & 65.240 & 99.434 & 98.996 \\
        \rowcolor{adaptedhighlight}
        \method{} (video-adapted)
        & 1.15 & \textbf{1.328} & \textbf{97.351} & \textbf{97.340} & 64.606
        & \textbf{65.301} & \textbf{99.439} & \textbf{99.004} \\
        \bottomrule
    \end{tabularx}
\end{table*}

\section{Conclusion}

We introduce \method{}, a KV-cache compression framework for autoregressive
video generation that predicts cache states across layers and quantizes the
remaining residual. Predictor parameters calibrated once offline can be shared
across videos for an efficient default, while optional per-block adaptation
trades additional memory and fitting time for higher fidelity. On HY-WorldPlay
and MAGI-1, both configurations improve fidelity over matched clustering
baselines while requiring less stored cache state, showing that better reconstruction need
not require a larger cache. Sharing these parameters also amortizes their cost
across concurrent video caches, so total cache memory grows more slowly with
video count. Together, these results provide a direction toward high-fidelity video
generation with smaller, more scalable KV caches.

\section{Limitations}
\label{sec:limitations}

We evaluate two autoregressive video models and one held-out prompt source;
gains vary with cache geometry, and we do not formally explain why linear
cross-layer prediction works or establish that it transfers to other models,
prompt distributions, or longer horizons. Memory results cover stored cache
state rather than whole-process peak GPU memory. The 16-video totals use
shared-parameter accounting validated with up to four retained videos
(Appendix~\ref{sec:appendix-concurrency}), not measured concurrent throughput;
generation latency was measured on MI355X. \method{} retains every historical
token position, leaving integration with token selection open. We interpret
no-reference VBench scores alongside paired BF16 fidelity and motion because
smoothness scores alone can reward motion suppression.

\section*{Acknowledgments}
We thank our colleagues at AMD who provisioned and maintained the compute
infrastructure used for these experiments.\label{sec:main-text-end}

\subsubsection*{AI use statement}
In this work, we used AI to help refine hypotheses and conceptual
framing; provide feedback on experimental methodology; assist with
implementation and debugging; organize experimental outputs; and support
qualitative and quantitative interpretation of results. We additionally used
it for brainstorming, literature discovery, citation verification, figure and
table preparation, and drafting and editing parts of the manuscript. We did
not use generative AI to generate evaluation datasets, formulate or prove
mathematical claims, or independently determine scientific conclusions. The
authors reviewed all AI-assisted work, tested code and numerical outputs,
checked reported results against evaluation outputs, and verified citations
against primary sources. We take responsibility for the final content of this
work, including text, claims, code, figures, and artifacts produced with the
aid of generative AI.

\subsubsection*{Ethics statement}
This work studies compression of internal state in existing video-generation
models and does not introduce human-subject data or train a new generative
model. The method inherits the potential biases, misuse risks, and content
limitations of the underlying generators; reducing memory use and inference
cost does not mitigate those risks and may make such systems easier to use.
Our evaluation concerns reconstruction fidelity, memory, and latency rather
than the safety or factual reliability of generated content. Users should
therefore retain the safeguards and usage policies of the underlying models.

\subsubsection*{Reproducibility statement}
Calibration, ablation, and 355-prompt evaluation splits are disjoint. Every
comparison uses matching model revisions, prompts, seeds, and conditioning
inputs. We treat prompts as the statistical units and form confidence intervals
with paired prompt-level bootstrap resampling. The appendix reports
cache-memory accounting and the latency protocol. Code and evaluation
data will be released.

\bibliography{coral-kv}
\bibliographystyle{unsrtnat}

\clearpage
\appendix

\section{Expanded related work}
\label{sec:appendix-related-work}

\paragraph{Autoregressive video generation.}
Causal and diffusion-forcing formulations generate video in temporal blocks
and retain prior context~\citep{chen2024diffusionforcing,yin2025causvid,
huang2025selfforcing}. Long-form and interactive systems include MAGI-1,
LongLive, LongLive-2.0, SkyReels-V2, LongCat-Video, and Matrix-Game 2.0
\citep{sandai2025magi1,yang2025longlive,chen2026longlive2,
chen2025skyreelsv2,meituan2025longcatvideo,he2025matrixgame2}; related backbones
and explicit-memory systems include Wan, HunyuanVideo, HunyuanVideo 1.5, and
WORLDMEM~\citep{wan2025,kong2024hunyuanvideo,hunyuan2025hunyuanvideo15,
xiao2025worldmem}.

\paragraph{Video-cache compression and attention reduction.}
QVG directly quantizes autoregressive video K/V with online codebooks, and
Quantized Keys Steal Attention studies the associated attention bias
\citep{xi2026qvg,tuncer2026qksa}. Forcing-KV and DensityKV select retained
cache content; PackCache compacts tokens; FastAR combines temporal compression
with sparse attention; VideoMLA and HeadCast alter the stored attention state;
and ISPA absorbs history into instance-specific parameters
\citep{ji2026forcingkv,zhao2026densitykv,li2026packcache,samuel2026fastar,
yesiltepe2026videomla,shen2026headcast,fu2026ispa}. ARCache instead reuses
denoising features~\citep{nan2026arcache}. Sparse VideoGen, Sparse VideoGen2,
and Radial Attention reduce attention computation and are complementary to
value compression~\citep{xi2025svg,yang2025svg2,li2025radialattention}.
Low-bit video-LLM cache quantization provides another video-domain analogue
\citep{tao2025vidkv}.

\paragraph{LLM KV compression and cross-layer structure.}
KIVI, KVQuant, ZipCache, and GEAR quantize or correct LLM cache state, while
TurboQuant and QJL develop online projection/vector-quantization schemes
\citep{liu2024kivi,hooper2024kvquant,he2024zipcache,kang2024gear,
zandieh2026turboquant,zandieh2025qjl}. AQUA-KV predicts the next layer and
quantizes its residual~\citep{shutova2025aquakv}. Cross-Layer Attention,
MiniCache, xKV, XQuant, DepthWeave-KV, FusedKV, and EchoKV instead share,
align, factorize, or reconstruct state across depth
\citep{brandon2024cla,liu2024minicache,chang2026xkv,yang2025xquant,
cordoba2026depthweave,lin2026fusedkv,ji2026echokv}. JoLT jointly allocates
Tucker ranks and codes a rotated residual~\citep{krishnan2026jolt}. Architectural
approaches such as YOCO, YOCO++, and multi-head latent attention reduce the
cache by design~\citep{sun2024yoco,wu2026yocopp,deepseekai2024deepseekv2};
token-selection approaches such as H2O and StreamingLLM are orthogonal
\citep{zhang2023h2o,xiao2024streamingllm}.

\paragraph{Quantization and residual coding.}
GPTQ is an influential post-training LLM quantization method that greedily corrects weight quantization error through successive least squares updates; Qronos further corrects not only weight quantization error, but also quantization error propagated through activations of previously quantized layers~\citep{frantar2023gptq,zhang2026qronos}.
QuIP, QuIP\#,
QuaRot, SpinQuant, ReSpinQuant, and PeRQ use incoherence processing, learned rotations, or permutations to
control outliers~\citep{frantar2023gptq,chee2023quip,tseng2024quipsharp,
ashkboos2024quarot,liu2025spinquant,kim2026respinquant,sanjeet2026mixquant}. EDEN supplies the
Gaussian-tuned quantization grid used by HIGGS; later work studies scale selection, signed grids, nested lattices, and rate--distortion
coding~\citep{vargaftik2022eden,malinovskii2025higgs,amboage2026piso,colbert2026signed,savkin2025nestquant,
domb2026hyperquant}. Atom studies low-bit serving-oriented quantization
\citep{zhao2024atom}.

\paragraph{Evaluation protocols.}
Our prompt suite comes from MovieGen~\citep{polyak2024moviegen}. We report
SSIM and LPIPS fidelity~\citep{wang2004ssim,zhang2018lpips}, use DINO features
for boundary analysis~\citep{caron2021dino}, and use dimensions from VBench
\citep{huang2024vbench}. VBench++, FVD, and DINOv2 provide related evaluation
and representation-learning context~\citep{huang2025vbenchpp,
unterthiner2018fvd,oquab2024dinov2}.

\section{Method and implementation details}

\subsection{Full quantization and reconstruction specification}
\label{sec:appendix-quantizer-specification}

We index the cache by patch tokens and define $N$ contiguous positions as a
\emph{cache block}; in our implementations, a cache block contains one or more
complete latent frames.  Completed blocks are encoded once and retained in
compressed form, while entries that are not yet eligible remain at native
precision.  For $K_\ell,V_\ell\in\mathbb{R}^{N\times C}$, layer zero is encoded
and reconstructed as
\begin{align}
Z^K_0&=\operatorname{Enc}_{\mathrm A}(K_0),&
Z^V_0&=\operatorname{Enc}_{\mathrm A}(V_0),
\label{eq:coral-anchor}\\
\widehat K_0&=\operatorname{Dec}_{\mathrm A}(Z^K_0),&
\widehat V_0&=\operatorname{Dec}_{\mathrm A}(Z^V_0).
\label{eq:coral-anchor-reconstruction}
\end{align}
For every $\ell>0$, the layer-specific predictors produce
\begin{align}
\widetilde K_\ell&=P^K_\ell(\widehat K_{\ell-1}),&
E^K_\ell&=K_\ell-\widetilde K_\ell,\label{eq:coral-key}\\
\widetilde V_\ell&=P^V_\ell([\widehat K_\ell,\widehat V_{\ell-1}]),&
E^V_\ell&=V_\ell-\widetilde V_\ell,\label{eq:coral-value}
\end{align}
where brackets concatenate channels. The residual codes and reconstructions
are
\begin{align}
Z^K_\ell&=\operatorname{Enc}_{\mathrm R}(E^K_\ell),&
Z^V_\ell&=\operatorname{Enc}_{\mathrm R}(E^V_\ell),
\label{eq:coral-residual-code}\\
\widehat K_\ell&=\widetilde K_\ell+
\operatorname{Dec}_{\mathrm R}(Z^K_\ell),&
\widehat V_\ell&=\widetilde V_\ell+
\operatorname{Dec}_{\mathrm R}(Z^V_\ell).
\label{eq:coral-reconstruction}
\end{align}
Reconstruction proceeds in layer order, with keys preceding values because
$P^V_\ell$ depends on $\widehat K_\ell$.  During encoding, reconstructed rather
than target tensors feed the next predictor, so encoder and decoder inputs
match.  The cache retains only $Z^K_0,Z^V_0$ and
$\{Z^K_\ell,Z^V_\ell\}_{\ell=1}^{L-1}$ with their scales and decoder metadata.

For completeness, the anchor's per-channel affine scalar quantizer maps a
value $x$ in channel $c$ to
\begin{equation}
z=\operatorname{clip}\!\left(
\operatorname{round}\!\left(\frac{x-\beta_c}{\alpha_c}\right),0,2^b-1\right),
\qquad \widehat x=\alpha_c z+\beta_c,
\label{eq:affine-quantizer}
\end{equation}
with FP16 scale and offset.  At low bitwidth, independent scalar quantization
can be dominated by outliers or energy concentrated in a few coordinates~\citep{sanjeet2026mixquant}.
Randomized orthogonal rotations can reduce coordinate
outliers~\citep{vargaftik2022eden,tseng2024quipsharp}. Hadamard-rotated LLM
weights and activations have been observed to approach Gaussian
distributions~\citep{malinovskii2025higgs,egiazarian2026mrgptq}; related
rotations have also been used for KV caches~\citep{ashkboos2024quarot}. A fixed
Hadamard transform has no such input-independent guarantee, so we select its
width empirically.

We distinguish scale-group size $g$ from transform size $h\mid g$.  For
$u\in\mathbb{R}^g$, store
$s=\lVert u\rVert_2/\sqrt{g/h}$, partition $u$ into $g/h$ blocks
$u_i\in\mathbb{R}^h$, and compute
\begin{equation}
y_i=H_h(u_i/s),\qquad H_hH_h^\top=hI.
\label{eq:hadamard-transform}
\end{equation}
Partitioning transformed coordinates into $p$-vectors $y_j$, assignment to the
nearest entry in a fixed codebook $\mathcal C=\{c_1,\ldots,c_M\}$ is
\begin{equation}
z_j=\arg\min_{m\in\{1,\ldots,M\}}\lVert y_j-c_m\rVert_2^2,
\label{eq:vector-quantizer}
\end{equation}
and decoding concatenates the selected vectors and applies
\begin{equation}
\widehat u_i=\frac{s}{h}H_h^\top\widehat y_i.
\label{eq:hadamard-inverse}
\end{equation}
With a $b_s$-bit stored scale, the storage cost is
$\log_2(M)/p+b_s/g$ bits/value before metadata. We use the published EDEN
codebook with $p=2$, $M=16$, and $h=64$. The selected key format uses
$(b_s,g)=(8,64)$ and the value format uses $(16,128)$, giving 2.125 bits/value
for either keys or values. The uniform FP16-per-64 control uses 2.25 bits/value.
Unlike analyses that prepend random signs, we use the same fixed block
Hadamard matrix throughout the channel dimension. We therefore evaluate its
distributional effect and select the transform width empirically in
Appendix~\ref{sec:appendix-residual-quantization}.

\paragraph{Model geometry and offline fitting.}
The effective predictor geometry is $(L,H_{\mathrm{KV}},d,C)=(30,24,256,6144)$
for HY-WorldPlay and $(34,8,128,1024)$ for MAGI-1. Offline calibration pools
all 384,000 and 1,024,000 cached token positions, respectively, from the 32
calibration rollouts. We fit each layer's block-diagonal term $D_\ell$ and
low-rank factors $A_\ell,B_\ell$ with three rounds of alternating headwise
ridge and reduced-rank regression updates, using regularization $10^{-3}$ and
seed zero. Normal equations are
accumulated in FP64 and solved by Cholesky, while the inputs and stored
predictor factors are BF16. On one MI355X, the complete fitting procedure took
24.8 minutes for HY-WorldPlay and 29.6 minutes for
MAGI-1, including data loading, sequential reconstruction, auditing, and
checkpointing but excluding generation of the calibration rollouts. The
stored predictors occupy 502.5 and 69.3 MB and are included in reported cache
memory.

We use three fitting rounds for all reported predictors. In a HY-WorldPlay
screen with disjoint calibration and evaluation cache dumps, six and ten rounds
reduced held-out KV relative MSE by 4.97\% and 7.55\% relative to three rounds.
The ten-round predictor improved PSNR by 0.10~dB on 32 selection-set videos
(95\% CI $[-0.21,0.41]$); its LPIPS and BF16-relative temporal differences
also spanned zero. We therefore retained three rounds for the reported method.

\paragraph{Video-specific adaptation.}
The video-adapted configuration retains the reusable rank-128 low-rank factors $A_\ell$ and
$B_\ell$. For each eligible cache block, it fits the block-diagonal term
$D_\ell$ and the intercept by independent per-head ridge solves over every
token in the block, again with regularization
$10^{-3}$. It then fits one least-squares scale and bias per output channel.
The solves use FP32 and the returned block, bias, and affine tensors use the
native BF16 cache dtype. These parameters are included in memory while their
block remains in the active attention history.

\paragraph{Stored residual representation.}
Each 64-value residual group stores 32 four-bit EDEN-2 indices, because each
index reconstructs a pair of transformed values. Key groups store one FP8
norm per 64 values, while value groups store one FP16 norm per 128 values.
The key and value residuals therefore each require $2.125$ bits/value before
alignment and container metadata. Layer zero uses stored four-bit affine codes
with one FP16 scale and offset per channel. Reported memory is obtained from
the allocated tensors, not from this bit-count calculation. The fixed 16-entry
EDEN-2 codebook is shared; its bytes are included once when it is stored
rather than embedded as a compile-time constant. We use the published
$d=2,n=16$ codebook without fitting centroids to our data. The video-adapted configuration and the
uniform-scale ablation instead use one FP16 scale per 64 residual values,
for a 2.25-bpv residual format.

\subsection{Efficient inference implementation}
\label{sec:appendix-efficient-inference}

A direct implementation of Eqs.~\ref{eq:coral-key}--\ref{eq:coral-reconstruction}
would decode the stored values, form two predictions per layer, add the residuals, and
rearrange the result on every cache read. Offline calibration removes predictor
fitting; the following optimizations reduce the remaining conversion overhead
without changing the representation or reconstruction rule.

\paragraph{Stored representation and factorized prediction.}
Stored cache blocks contain only anchor or residual codes, group scales,
and decoding metadata. Reported memory includes one shared copy of the predictors
and codebook; reconstructed K/V and workspaces are transient. The block-diagonal
term uses batched per-head contractions and the low-rank factors use thin
matrix multiplications, never materializing $D_\ell+A_\ell B_\ell$. The
implementation reuses prepared casts and eligible precomputed projections, accepts
strided head blocks directly, and accumulates components in place.
Synchronized profiles give dense-to-block-DPLR predictor-application ratios of
$2.64\times$ on HY-WorldPlay and $0.76\times$ on MAGI-1. We therefore select
the factorized form for memory efficiency rather than assuming
universal speedup, and use separate unprofiled runs in both execution orders for end-to-end
latency claims.

\paragraph{Fused compressed-cache conversion.}
Where model shapes permit, GPU kernels combine the transform, codebook assignment,
encoding, decoding, residual addition, and attention-layout operations. The decoder
combines the prediction with the decoded $H_{64}$/EDEN residual in place and emits
the attention-facing layout without retaining an expanded cache block, avoiding
intermediate allocations and redundant casts, transposes, and copies. Entries
that will not be read again are skipped. HY-WorldPlay uses approved fused
assignment and reconstruction kernels. MAGI-1 uses fused codebook
assignment while retaining conservative transform and reconstruction dispatch
where broader fusion does not pass the numerical checks. Both implement the same
stored representation and reconstruction rule.

\paragraph{Matched attention implementation and numerical checks.}
All methods within a model use the same attention implementation: AITER for HY-WorldPlay and the
native grouped-query attention implementation for MAGI-1. HY-WorldPlay is checked at the
reconstructed-tensor and final-BF16-output levels; MAGI-1 additionally uses a
paired decoded-video check. Only implementations passing these checks enter the latency
study, so differences do not arise from a relaxed algorithm or attention
implementation.

\section{Experimental protocol}

\subsection{Evaluation configurations}
\label{sec:appendix-evaluation-configurations}

\begin{table*}[t]
    \centering
    \caption{Evaluation configurations. Attention implementations are held fixed
    across methods within each model.}
    \label{tab:model-setup}
    \scriptsize
    \setlength{\tabcolsep}{7pt}
    \begin{tabular*}{\textwidth}{@{\extracolsep{\fill}}lccccc@{}}
        \toprule
        Model
        & Resolution
        & Frames
        & Denoising steps
        & \shortstack{First compression-\\dependent frame}
        & \shortstack{Attention\\implementation} \\
        \midrule
        HY-WorldPlay
        & $704\times1280$ & 285 & 4 & 29 & AITER \\
        MAGI-1
        & $720\times720$ & 192 & 16 & 24 & native GQA \\
        \bottomrule
    \end{tabular*}
\end{table*}

\begin{table}[h]
    \centering
    \caption{Disjoint prompt splits and their permitted uses.}
    \label{tab:split-contract}
    \small
    \begin{tabular}{lcl}
        \toprule
        Split & Prompts & Permitted use \\
        \midrule
        Calibration & 32 & Fit offline predictors and quantizer parameters \\
        Selection & 32 & Select configurations and report ablations \\
        Evaluation & 355 & Held-out comparisons and confidence intervals \\
        \bottomrule
    \end{tabular}
\end{table}

The split membership is immutable and prompt-length stratified. A comparison
is admitted only when model revision, prompt index, seed, conditioning image,
generation geometry, and scorer configuration match. HY-WorldPlay methods
also share the same action sequence. Offline predictors are fitted only on the
calibration split; neither selection nor evaluation cache tensors are used
to update them. Configuration choices are fixed after selection-set analysis
and before examining the 355-prompt evaluation aggregates.

\begin{figure}[t]
    \centering
    \includegraphics[width=0.96\textwidth]{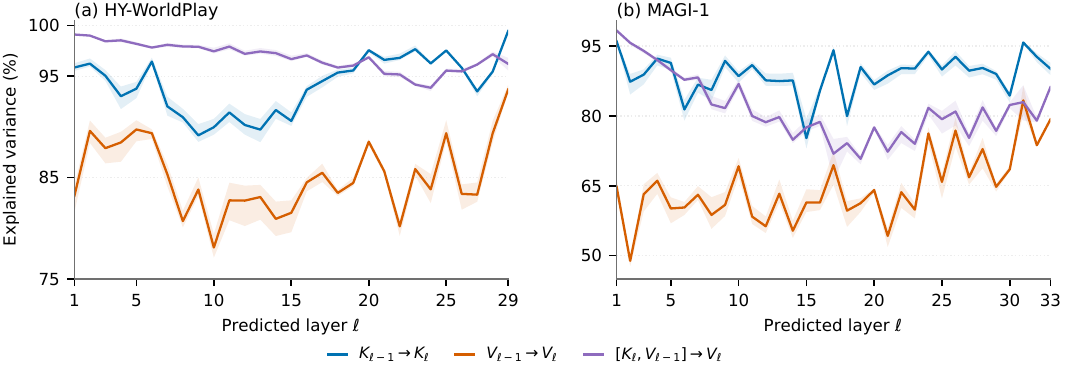}
    \caption{\textbf{Cross-layer prediction across depth.} The full
    layerwise measurements summarized in Figure~\ref{fig:motivation}(a).
    Lines and shading show the mean and range over three held-out cache dumps.}
    \label{fig:layerwise-predictability}
\end{figure}

\paragraph{Layerwise cross-layer-predictability probe.}
Figure~\ref{fig:motivation}(a) and Appendix
Figure~\ref{fig:layerwise-predictability} evaluate full affine ridge predictors on
unquantized KV states, using 4,096 evenly spaced token positions per cache dump
and a ridge rate of $10^{-3}$. Figure~\ref{fig:motivation}(a) averages over
predicted layers and three rollout-held-out evaluations; the appendix curves
show layerwise means and ranges across those evaluations. For HY-WorldPlay, the key and
preceding-value curves fit each held-out rollout from the other 15 rollouts;
the combined-input value curve evaluates a fixed full-affine map calibrated
on eight disjoint rollouts. For MAGI-1, each of the first three held-out
rollouts is predicted by maps fit on the other 15 rollouts. We report
$1-\operatorname{mean}_c\operatorname{Var}_t(\widehat y-y)_c/
\operatorname{mean}_c\operatorname{Var}_t(y)_c$ at each predicted layer.

\paragraph{Reconstruction at matched parameter storage.}
\label{sec:appendix-equal-parameter-storage}
Figure~\ref{fig:motivation}(b) compares rank-128 block-DPLR predictors
with codebooks containing 705 centroids per head, role, and layer on
HY-WorldPlay and 513 on MAGI-1. Predictor weights and centroid vectors each
occupy 502,456,320 BF16 bytes on HY-WorldPlay and 69,341,184 on MAGI-1.
Assignment IDs require
additional storage for the centroid methods. Predictors and offline-trained
centroids use the same 32 calibration videos; online centroids are fitted to
each held-out target block. Each arm reconstructs the same KV tokens from
eight held-out videos per model. Predictors receive true BF16 preceding-layer
keys and values; the value predictor receives the predicted current-layer
key, not the true key. No residual codes, scales, or INT4 anchors are used.
Relative MSE divides squared reconstruction error by the centered squared sum
of the target, first aggregating across predicted layers within each video and
then averaging over videos. Table~\ref{tab:equal-parameter-reconstruction}
gives the relative-MSE values underlying Figure~\ref{fig:motivation}(b).
Predictors have lower relative MSE than both centroid variants on all eight
paired videos in each model and KV role. This comparison evaluates
conditional reconstruction with clean predecessor states; it does not measure
recursive reconstruction through a generated video.

\begin{table}[t]
\centering
\caption{\textbf{Relative MSE at equal BF16 parameter storage.} All arms
reconstruct the same held-out KV tokens before residual quantization. Lower
is better. Centroid assignments require additional storage.}
\label{tab:equal-parameter-reconstruction}
\small
\begin{tabular}{llrrr}
\toprule
Model & State & Predictor & Offline centroids & Online centroids \\
\midrule
HY-WorldPlay & Keys & \textbf{0.0669} & 0.2522 & 0.1626 \\
             & Values & \textbf{0.2618} & 0.6403 & 0.4445 \\
MAGI-1       & Keys & \textbf{0.0080} & 0.0270 & 0.0162 \\
             & Values & \textbf{0.3806} & 0.6959 & 0.5226 \\
\bottomrule
\end{tabular}
\end{table}
\FloatBarrier

\section{Additional metric and statistical details}

\subsection{QBD implementation and robustness}
\label{sec:appendix-qbd-definition}

The first affected decoded-frame indices in Eq.~\ref{eq:qbd} follow the known
cache-quantization schedule and VAE temporal decoding. The main-text
definition takes the lowest similarity over three transitions and excludes
comparison locations within three frames of any first affected frame
or at the video ends. To test sensitivity, we take the lowest similarity over
two, three, or four consecutive transitions and exclude comparison-window
starts within two to five frames of each first affected frame, yielding 12
definitions.
For prompt-level comparisons, we normalize each compressed video's QBD by its
paired BF16 video's QBD under the same boundary schedule. Both \method{} configurations have
lower BF16-normalized QBD than QVG-Pro on both models across all 12 definitions.
Under the primary definition, prompt-paired 95\% bootstrap intervals for the
BF16-normalized QBD advantage over QVG-Pro exclude zero for both configurations
on both models.

\subsection{VBench and BF16-relative dynamics}
\label{sec:appendix-vbench-dynamics}

\paragraph{VBench-6 diagnostics.}
We complement reference-based fidelity with a compression-focused subset of
six official VBench dimensions~\citep{huang2024vbench}: subject consistency,
background consistency, imaging quality, aesthetic quality, motion smoothness,
and temporal flickering. We call this subset VBench-6 and report all six
coordinates separately as absolute scores; it is not a new aggregate
benchmark. Subject and background consistency
retain the clean first frame as a global reference, while the other dimensions
operate on the contiguous post-quantization window. For HY-WorldPlay,
full-reference fidelity uses frames 30--284, while QBD and VBench-6 use frames
29--284. For MAGI-1, VBench-6 uses frames 24--95 so that it remains comparable
to the conventional 96-frame setting; full-reference fidelity uses frames
25--191 and QBD uses frames 24--191. Since no-reference motion
smoothness and flicker can reward suppressed motion, we also compare temporal
dynamics with BF16 before attributing a gain to improved stability.

\paragraph{BF16-relative temporal dynamics.}
To distinguish genuine stability from motion suppression, we area-downsample
the post-quantization window to $64\times36$ grayscale frames and measure its
mean absolute inter-frame difference and spatial total variation.  We divide
both quantities by the value from the paired BF16 video; ratios near one
preserve the reference motion magnitude and spatial variation.  We also
measure the absolute difference between the compressed and BF16 frame-to-frame
residuals, normalized by the BF16 inter-frame difference, for which zero is
ideal. These quantities are diagnostics rather than replacement quality
metrics.

\subsection{Per-boundary QBD}
\label{sec:appendix-qbd-boundaries}

\begin{figure*}[t]
    \centering
    \includegraphics[width=0.96\textwidth]{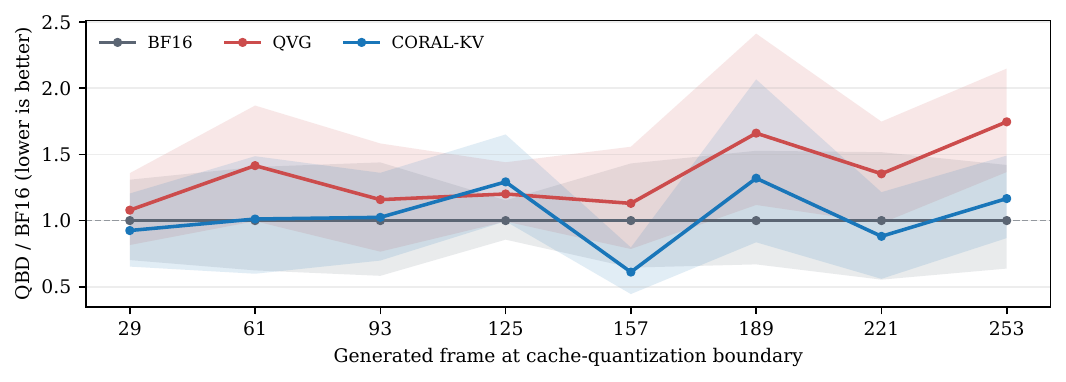}
    \caption{\textbf{Boundary stability over long-horizon generation.}
    BF16-normalized QBD at all eight cache-quantization boundaries on the
    32-prompt HY-WorldPlay selection set. Lines are prompt means and bands
    are 95\% bootstrap intervals. Lower is better.}
    \label{fig:qbd-per-boundary}
\end{figure*}

Aggregate QBD can hide whether discontinuities worsen as generation proceeds.
Figure~\ref{fig:qbd-per-boundary} therefore resolves all eight scheduled
HY-WorldPlay boundaries on the 32-prompt selection set. Across the complete
285-frame rollout, mean QBD is $1.035\times$ BF16 for \method{} and
$1.321\times$ for QVG-4; the paired QVG-minus-\method{} difference in raw QBD is
$0.472$ ($95\%$ CI $[0.242,0.702]$), with improvements on 26 of 32 prompts.
At the final boundary, the corresponding ratios are $1.166\times$ and
$1.747\times$. QVG's fitted QBD slope is positive at $0.149$ per boundary
($[0.052,0.253]$), whereas \method{} has a smaller slope of $0.053$ whose
interval includes zero ($[-0.053,0.161]$).

\section{Comparison with token selection}
\label{sec:appendix-token-selection}

Token selection reduces memory by discarding historical positions rather than
compressing their keys and values. We port the released DensityKV
selection rule~\citep{zhao2026densitykv} to both video generators and also
compare with a control that keeps the most recent historical BF16 tokens.
All methods use the same local window, first-frame tokens, generation settings,
and paired BF16 references. For each \method{} variant, we cap each selector's
K/V and metadata at the same compressed-history memory. On HY-WorldPlay, this
active-window budget excludes the common 5.19-GB BF16 suffix;
Table~\ref{tab:main-quality-memory} instead reports full-history memory.

We use DensityKV's released per-head selection rule based on post-RoPE key
density. On HY-WorldPlay, it processes each completed four-frame block.
MAGI-1's 12,150-token blocks exceed the matched retained-token budget, so we
initialize the retained set with uniformly spaced positions from the first block and process later
blocks in chronological 512-token groups. When MAGI-1 rewrites a block, we
refresh K/V for retained positions without changing their density metadata.
These changes accommodate the models' update schedules; the selection rule is
unchanged.

\begin{table*}[t]
    \centering
    \caption{\textbf{\method{} has higher BF16-relative PSNR than both token-selection
    controls at matched cache memory.} Mean RGB PSNR (dB; higher is better) on
    16 HY-WorldPlay and 64 MAGI-1 paired prompts. Memory is the
    retained compressed-history budget; each control's K/V and any selection metadata
    fit within it. These cohorts are separate from the 355-prompt
    evaluation in Table~\ref{tab:main-quality-memory}.}
    \label{tab:token-selection}
    \small
    \setlength{\tabcolsep}{6pt}
    \begin{tabular*}{\textwidth}{@{\extracolsep{\fill}}llrrrr@{}}
        \toprule
        Model (prompts) & Variant & Memory (GB) & \method{} & DensityKV & Recent BF16 \\
        \midrule
        HY-WorldPlay (16) & Reusable & 4.05 & 16.47 & 11.76 & 11.52 \\
        HY-WorldPlay (16) & Video-adapted & 4.52 & 16.78 & 11.67 & 11.75 \\
        MAGI-1 (64) & Reusable & 0.99 & 31.45 & 28.18 & 27.31 \\
        MAGI-1 (64) & Video-adapted & 1.15 & 32.97 & 28.67 & 27.38 \\
        \bottomrule
    \end{tabular*}
\end{table*}

We score HY-WorldPlay frames 30--284 on 16 paired MovieGen prompts and MAGI-1
frames 25--191 on 64 paired prompts, averaging frame PSNR within each prompt
before averaging prompts. At matched memory, \method{} exceeds DensityKV by
3.27--5.11~dB and the recent-token control in every row
(Table~\ref{tab:token-selection}). All four paired 95\% prompt-bootstrap
intervals for the DensityKV gains exclude zero (smallest lower bound 2.76~dB);
MAGI-1 SSIM and LPIPS agree.
Thus, in these settings, compressing all historical positions preserves more
BF16 fidelity than retaining selected BF16 positions. A disjoint eight-prompt
HY-WorldPlay follow-up gives the same PSNR ordering. The HY-WorldPlay comparison
is exploratory because BF16 outputs vary across processes; each method is
compared with its own paired BF16 reference. These tests measure fidelity at
matched cache memory, not perceptual quality on DensityKV's original backbones.

\section{Memory accounting and implementation controls}

For method $m$, reported compressed-history memory is
\begin{equation}
    M_m = M_{\mathrm{codes}} + M_{\mathrm{scales}} +
          M_{\mathrm{metadata}} + M_{\mathrm{pred}} + M_{\mathrm{online}},
    \label{eq:appendix-memory}
\end{equation}
where the cache-block terms cover all nine completed HY-WorldPlay blocks or
MAGI-1's retained attention window, and $M_{\mathrm{pred}}$ contains one complete
set of offline predictors, and
$M_{\mathrm{online}}$ contains block-specific parameters required for future
reconstruction. QVG codebooks are included for every encoded block that needs
them. We do not divide reusable predictor memory by the number of prompts or
videos in the experiment. HY-WorldPlay totals use the checkpoint after nine
complete blocks, so no incomplete native-precision suffix is present. The
fidelity evaluation instead retains 48 latent frames, as described in
Section~\ref{sec:experimental-methodology}.
Reconstructed tensors, attention outputs, and conversion workspaces are
temporary and excluded; these analytical cache-memory totals are not
peak process-memory measurements.
MAGI-1 retains at most four preceding
24-frame chunks (96 previous frames); older blocks and their associated video-specific
parameters are excluded for every method because no future query can attend them.
For video-adapted CORAL-KV, the current block-diagonal parameters belong to and
are stored with the newest compressed block and are counted only as part of that block.

\subsection{Full-history cache scaling}
\label{sec:appendix-full-history-memory}

Table~\ref{tab:main-quality-memory} uses a full-history HY-WorldPlay cache that retains every
completed 7,040-token block. The 285-frame generation contains nine such
blocks (72 latent frames). Each table entry is the total stored KV state
immediately after the listed blocks have been compressed, so no incomplete
BF16 block is present. QVG retains its codebooks and assignments for every
block; \method{} retains the encoded blocks and includes its 0.502-GB predictor
parameters once. The per-block byte counts are taken from the packed-cache
measurements before the 48-latent-frame context begins replacing older blocks.
The measured fidelity results use that retained context.

\begin{table}[h]
    \centering
    \caption{\textbf{Reusable \method{} reduces the memory cost of retaining
    full video history.} Each block contains eight latent frames, or 7,040
    visual tokens; nine blocks cover the 285-frame HY-WorldPlay video.
    The totals include QVG codebooks or the reusable predictor parameters.}
    \label{tab:full-history-memory}
    \small
    \setlength{\tabcolsep}{5pt}
    \begin{tabular}{lrrr}
        \toprule
        & \multicolumn{3}{c}{Total memory (GB)} \\
        \cmidrule(lr){2-4}
        Method & 1 block & 5 blocks & 9 blocks \\
        \midrule
        QVG-1 & \textbf{0.89} & 4.44 & 7.99 \\
        QVG-4 & 1.48 & 7.42 & 13.36 \\
        QVG-Pro & 1.61 & 8.03 & 14.46 \\
        \rowcolor{reusablehighlight}
        \method{} (reusable) & 1.21 & \textbf{4.06} & \textbf{6.90} \\
        \rowcolor{adaptedhighlight}
        \method{} (video-adapted) & 1.53 & 5.63 & 9.74 \\
        \bottomrule
    \end{tabular}
\end{table}

\subsection{Memory sharing across concurrent videos}
\label{sec:appendix-concurrency}

Table~\ref{tab:main-quality-memory} includes one complete copy of the reusable
predictor parameters in the one-video total. In concurrent serving, those predictors are
shared while every video retains an independent compressed cache. For $N$ simultaneous
videos, the corresponding compressed-cache memory is therefore $B+NC$, where $B$ is the
shared predictor memory and $C$ is the per-video compressed-cache memory. QVG instead
fits and retains video-specific centroids, so its compressed-cache memory
scales as $NQ$.
The $N=1$ and $N=16$ columns of Table~\ref{tab:main-quality-memory} report
$B+C$ and $B+16C$, respectively, using nine blocks per HY-WorldPlay video.

\subsection{Matched latency definition and results}
\label{sec:appendix-latency}

We measure latency overhead in a dedicated matched experiment with identical
prompts, seeds, hardware allocation, and attention implementations. For method $m$,
we report its additional generation time over the local BF16 reference,
\begin{equation}
    O_m = \frac{\sum_{p\in\mathcal{P}} T_m(p)}
                 {\sum_{p\in\mathcal{P}} T_{\mathrm{BF16}}(p)}-1,
    \label{eq:latency-overhead}
\end{equation}
where $\mathcal{P}$ is the matched prompt set. We average $O_m$ over two execution
orders and report it as a percentage. Timing includes online fitting and excludes
compilation, warm-up, and video export. Table~\ref{tab:main-quality-memory}
reports overheads from separate matched experiments. In the direct QVG-Pro
comparison, reusable \method{} adds $10.7\%$ versus $17.7\%$ on HY-WorldPlay
and $6.8\%$ versus $9.6\%$ on MAGI-1. Both execution orders favor \method{},
although two runs do not establish a statistically significant speedup. A
separate matched comparison confirms that video adaptation increases overhead
relative to reusable \method{} on both models. We do not compare overheads
across separate experiments.

\section{Ablations and robustness}
\label{sec:appendix-robustness}

\subsection{Configuration selection}
\label{sec:appendix-configuration-selection}

We select the adapter structure and learning regime, residual transform and
codebook, and layer-zero representation once on the 32-prompt
selection set. It is disjoint from calibration and held-out evaluation;
all comparisons use matched prompts, seeds, and conditioning inputs and score
complete generated videos using reconstructed caches rather than cache-space regression alone.

For the no-predictor comparison in Figure~\ref{fig:predictor-selection}, the
rank-128 offline predictor raises PSNR from 14.14 to 17.55~dB on HY-WorldPlay
and from 25.19 to 31.99~dB on MAGI-1. The paired gains are 3.41~dB (95\% CI
$[2.76,4.12]$) and 6.81~dB ($[6.25,7.39]$), with 32/32 wins. All four paired
intervals for block-DPLR against block-diagonal-only and low-rank-only
predictors exclude zero. Across ranks 16 to 128, offline block-DPLR rises from
17.08 to 17.55~dB and from 28.84 to 31.99~dB; rank 256 leaves both means
unchanged to two decimals. Low-rank only remains within 14.39--14.61 and
25.37--25.53~dB. Video-specific block-diagonal fitting continues to rank 256,
reaching 17.89 and 33.85~dB, at higher memory and fitting cost.

We also tested which predictor component to refit online. In an exploratory
eight-video HY-WorldPlay diagnostic, refitting the rank-128 low-rank component
instead of the block-diagonal component changed PSNR by only $+0.10$~dB and
slightly worsened LPIPS (0.329 versus 0.327). In a separate rank-64 timing
profile, the global low-rank update took $2.6\times$ as long to fit as the
block-diagonal update, whose independent per-head solves can be parallelized.
These diagnostics predate the
final residual-quantization configuration and motivate the selected adaptation
path rather than serving as headline evidence.

The rank sweep intentionally varies predictor memory. To isolate structure at a
fixed budget, we additionally fit the largest standalone low-rank predictors
not exceeding block-DPLR r128. All other components and all 32 paired prompts
are fixed. Table~\ref{tab:state-matched-lowrank} shows that the standalone
controls use slightly fewer bytes, yet block-DPLR has a positive paired PSNR
interval on both models. SSIM and MSE also favor block-DPLR on both; HY LPIPS
and QBD intervals cross zero, while all five MAGI-1 metrics favor block-DPLR
with intervals excluding zero. Thus the gain is not explained by a larger
predictor-memory budget.

\begin{table}[t]
    \centering
    \caption{\textbf{Equal-memory structure control on the 32-prompt selection set.} Deltas are
    standalone low-rank minus block-DPLR r128; brackets are paired 95\% CIs.}
    \label{tab:state-matched-lowrank}
    \scriptsize
    \setlength{\tabcolsep}{2.6pt}
    \begin{tabular}{lcccc}
        \toprule
        Model & LR rank & LR/DPLR MB & $\Delta$PSNR (dB) & $\Delta$SSIM \\
        \midrule
        WorldPlay & 281 & 501.4/502.5
        & $-0.519$ $[-0.867,-0.179]$
        & $-0.0125$ $[-0.0247,-0.0005]$ \\
        MAGI-1 & 204 & 69.1/69.3
        & $-2.778$ $[-3.240,-2.344]$
        & $-0.0328$ $[-0.0435,-0.0235]$ \\
        \bottomrule
    \end{tabular}
\end{table}

\begin{figure*}[t]
    \centering
    \includegraphics[width=0.96\textwidth]{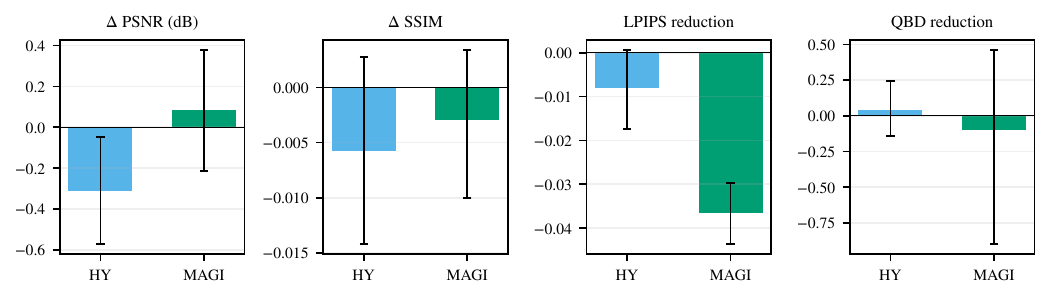}
    \caption{\textbf{Layer-zero anchor ablation on the 32-prompt selection set.} Bars show the paired
    effect of replacing the INT4 anchor with BF16; positive values favor BF16.
    Error bars are paired 95\% bootstrap confidence intervals over 32 prompts.}
    \label{fig:anchor-selection}
\end{figure*}

\paragraph{Layer-zero anchor.}
Holding the selected predictor and residual quantization fixed, BF16 increases cache
memory from 4.85 to 5.18 bits per BF16 cache value on HY-WorldPlay
and from 2.47 to 2.82 on MAGI-1. It changes PSNR by $-0.31$~dB on HY-WorldPlay
(95\% CI $[-0.57,-0.05]$) and $+0.09$~dB on MAGI-1 (CI $[-0.21,0.38]$),
while worsening MAGI-1 LPIPS by 0.036 (CI $[0.030,0.044]$). Thus BF16 adds
memory without a quality advantage; Figure~\ref{fig:anchor-selection} reports
the paired ablation.

\subsection{Residual transform and codebook}
\label{sec:appendix-residual-quantization}
\begin{figure*}[t]
    \centering
    \includegraphics[width=0.96\textwidth]{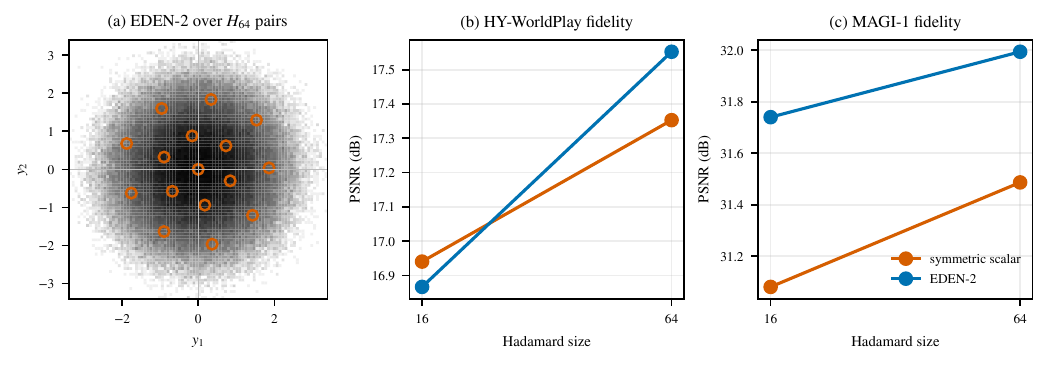}
    \caption{\textbf{EDEN-2 codebook and quantizer selection.}
    (a) The 16 EDEN-2 reconstruction points over representative
    $H_{64}$-transformed residual pairs from MAGI-1. (b--c) On the 32-prompt
    selection set, crossing $H_{16}$/$H_{64}$ with scalar/EDEN-2 assignment
    at a fixed 64-value scale group shows the highest PSNR for
    $H_{64}$/EDEN-2 on both models.}
    \label{fig:residual-selection}
\end{figure*}

We next hold the predictor and layer-zero anchor fixed and vary the residual
quantization. To measure generated-video quality, we hold the 64-value scale
group fixed and cross transform sizes
$h\in\{16,64\}$ with a symmetric scalar grid and the EDEN-2 grid. This
controlled $2\times2$ comparison uses the same bit allocation and separates the
effect of broader Hadamard mixing from vector rather than scalar assignment.
Figure~\ref{fig:residual-selection}(a) shows the 16 EDEN-2 reconstruction
points over representative $H_{64}$-transformed residual pairs, and
Figure~\ref{fig:residual-selection}(b,c) reports the resulting video fidelity.

We test sharing each
FP16 scale across 128 rather than 64 values. This reduces total cache memory by
only 2.1\% on HY-WorldPlay and 4.9\% on MAGI-1, while lowering PSNR
by 0.23 and 0.82~dB and increasing LPIPS by 0.0080 and 0.0094, respectively.
We therefore reject uniform FP16 groups of 128. We next evaluate three
predeclared 2.125-bpv assignments. Relative to the reversed allocation (K
FP16/128, V FP8/64), the selected K FP8/64, V FP16/128 format improves PSNR by
0.21/0.67~dB and reduces LPIPS by 0.0080/0.0063 on
HY-WorldPlay/MAGI-1. Relative to uniform FP16/128, it gains 0.08/0.85~dB and
reduces LPIPS by 0.0041/0.0087. These comparisons motivated the key/value-specific
scale allocation, which we then held fixed for held-out evaluation. Uniform FP16
groups of 64 are used by the video-adapted configuration and ablations.

\subsection{Online centroids, offline codebooks, and reusable predictors}
\label{sec:appendix-base-residual-grid}
We compare online QVG-1, QVG-1 with offline-trained codebooks, and reusable
\method{} predictors under two residual codecs. Online QVG-1 fits a 256-centroid
codebook per head, key or value, and predicted layer for each cache block. The
offline-codebook variant fits the same codebooks once on the 32-video
calibration set and reuses them across blocks and videos. \method{} uses one
reusable set of rank-128 block-DPLR predictors, calibrated separately for each
codec on the same videos. All three share a simulated INT4 layer-zero anchor.
Each uses either scalar INT2 with one FP8 scale per 64 residual values (2.125
bits/value) or $H_{64}$/EDEN-2 with one FP16 scale per 64 values (2.25
bits/value).
We evaluate eight paired held-out prompts on each model, generating 285
HY-WorldPlay frames and 96 MAGI-1 frames. PSNR is measured against each
prompt's BF16 generation after the first compressed block affects the video.

\begin{table}[t]
    \centering
    \caption{\textbf{At similar total compressed-state storage, reusable
    prediction improves PSNR over offline-trained codebooks with either
    residual quantizer.} Online QVG-1 fits
    codebooks for each cache block; the offline-codebook variant reuses
    calibration-fitted codebooks. Storage includes anchors, residual codes and scales,
    assignments, and codebooks or predictors. HY totals nine generated cache
    blocks, although fidelity generation retains the most recent 48 latent
    frames; MAGI-1 retains four 12,150-token blocks. Bold marks the highest
    PSNR within each model and residual quantizer.}
    \label{tab:base-residual-grid}
    \small
    \setlength{\tabcolsep}{5pt}
    \begin{tabular}{@{}llrr@{}}
        \toprule
        Method & Residual codec & PSNR $\uparrow$ & Storage (GB) $\downarrow$ \\
        \midrule
        \multicolumn{4}{@{}l}{\textbf{HY-WorldPlay (285 frames)}} \\
        QVG-1 (online) & Scalar INT2 & 15.941 & 8.117 \\
        QVG-1 (offline codebooks) & Scalar INT2 & 15.299 & 6.657 \\
        \method{} (reusable) & Scalar INT2 & \textbf{16.162} & 6.889 \\
        QVG-1 (online) & $H_{64}$/EDEN-2 & 16.249 & 8.470 \\
        QVG-1 (offline codebooks) & $H_{64}$/EDEN-2 & 14.910 & 7.010 \\
        \method{} (reusable) & $H_{64}$/EDEN-2 & \textbf{16.576} & 7.242 \\
        \midrule
        \multicolumn{4}{@{}l}{\textbf{MAGI-1 (96 frames)}} \\
        QVG-1 (online) & Scalar INT2 & 30.317 & 1.086 \\
        QVG-1 (offline codebooks) & Scalar INT2 & 30.148 & 0.982 \\
        \method{} (reusable) & Scalar INT2 & \textbf{30.505} & 0.992 \\
        QVG-1 (online) & $H_{64}$/EDEN-2 & 34.433 & 1.138 \\
        QVG-1 (offline codebooks) & $H_{64}$/EDEN-2 & 34.369 & 1.034 \\
        \method{} (reusable) & $H_{64}$/EDEN-2 & \textbf{38.100} & 1.043 \\
        \bottomrule
    \end{tabular}
\end{table}

The reusable predictor reduces mean first-block base-reconstruction relative
MSE by 36--53\% versus online QVG-1 and 56--70\% versus offline-trained
codebooks across the four matched-codec comparisons; it also lowers
final-cache relative MSE in each case. Relative to offline codebooks, it
improves PSNR by 0.86 and 1.67~dB on HY-WorldPlay under scalar INT2 and
$H_{64}$/EDEN-2, respectively, and by 0.36 and 3.73~dB on MAGI-1. The
MAGI-1 scalar interval includes zero; the other three paired 95\% intervals
exclude zero. Relative to online QVG-1, the gains are small on HY-WorldPlay
and under scalar INT2 on MAGI-1, but reach 3.67~dB under $H_{64}$/EDEN-2 on
MAGI-1. The predictor weights occupy 0.502/0.069~GB on HY-WorldPlay/MAGI-1,
versus 0.182/0.035~GB for offline codebooks; residual codes make their total
storage similar at the reported video lengths. The two residual codecs also
differ in scale precision, and each predictor is calibrated for its codec, so
their comparison does not isolate the Hadamard transform alone.

\clearpage
\subsection{Calibration-set size}
\label{sec:appendix-calibration-scaling}

\begin{table}[h]
    \centering
    \caption{\textbf{Effect of calibration-set size.} Rank-128 block-DPLR
    predictors fitted on 2, 8, 16, or 32 calibration prompts (one video per
    prompt) are evaluated on
    the same 32-prompt selection set. PSNR (dB) is from a separate
    calibration-size experiment; MAGI-1 uses 96 frames.}
    \label{tab:calibration-scaling}
    \small
    \setlength{\tabcolsep}{6pt}
    \begin{tabular}{lrrrr}
        \toprule
        Calibration prompts & 2 & 8 & 16 & 32 \\
        \midrule
        HY-WorldPlay & 17.58 & 18.02 & \textbf{18.59} & 17.94 \\
        MAGI-1 & 37.22 & 37.89 & 38.06 & \textbf{38.39} \\
        \bottomrule
    \end{tabular}
\end{table}

MAGI-1 improves across this sweep; HY-WorldPlay peaks at 16 calibration
prompts. We use 32 for both models, as specified before evaluating the sweep.
Because this experiment used separate video generations, its absolute PSNR
values should be compared within the table rather than with
Figure~\ref{fig:predictor-selection}.

\section{Additional qualitative results}

\subsection{Long-horizon robustness}
\label{sec:appendix-long-horizon}

Our main MAGI-1 evaluation generates 192 decoded frames for all 355 held-out
prompts. This covers the conventional 96-frame setting and a continuation of
equal length without changing prompts, seeds, conditioning inputs, or any
compression setting. We report the standard post-quantization window (frames
24--95), early continuation (96--143), and late continuation (144--191).
The selected reusable and video-adapted configurations are compared with
QVG-Pro at the prompt level using 95\% bootstrap confidence intervals. Before
the full evaluation, we verified the extended rollout on a fixed subset of the
selection set without using method outputs for selection.

\begin{figure*}[!htbp]
    \centering
    \includegraphics[width=\textwidth]{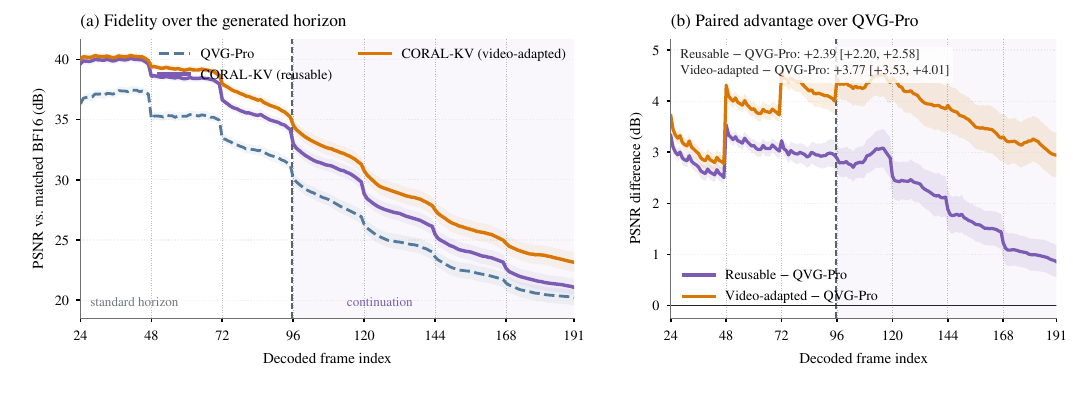}
    \caption{\textbf{MAGI-1 fidelity over 192 decoded frames.}
    Across all 355 held-out prompts, (a) prompt-mean PSNR and 95\% bootstrap
    bands are measured against matched BF16 generation for QVG-Pro and the two
    reported CORAL-KV configurations; (b) shows paired CORAL-minus-QVG-Pro
    PSNR differences and intervals. Dotted lines mark cache boundaries; the
    dashed line begins the continuation beyond the standard 96-frame horizon.
    Curves are unsmoothed.}
    \label{fig:magi-long-horizon}
\end{figure*}

Figure~\ref{fig:magi-long-horizon} shows that the predictor advantage persists
beyond the standard horizon for both reported CORAL-KV configurations. On
frames 24--95, reusable and video-adapted \method{} improve over QVG-Pro by 2.97~dB (95\% CI
$[2.85,3.10]$; 351/355 wins) and 3.74~dB ($[3.59,3.88]$; 354/355 wins).
The corresponding gains are 2.61 and 4.23~dB on frames 96--143, and remain
1.32~dB ($[1.02,1.63]$; 250/355 wins) and 3.36~dB
($[2.95,3.78]$; 297/355 wins) in the late frames 144--191. SSIM and LPIPS
agree in every window; in the late window, reusable/video-adapted \method{} improve SSIM by
0.0355/0.0675 and reduce LPIPS by 0.0345/0.0638. The frames 25--191
aggregate reproduces the main-table gains of 2.39 and 3.77~dB. The reusable
advantage narrows late in the rollout.

\subsection{Additional reconstruction examples}

\begin{figure}[!ht]
    \centering
    \includegraphics[width=0.80\textwidth]{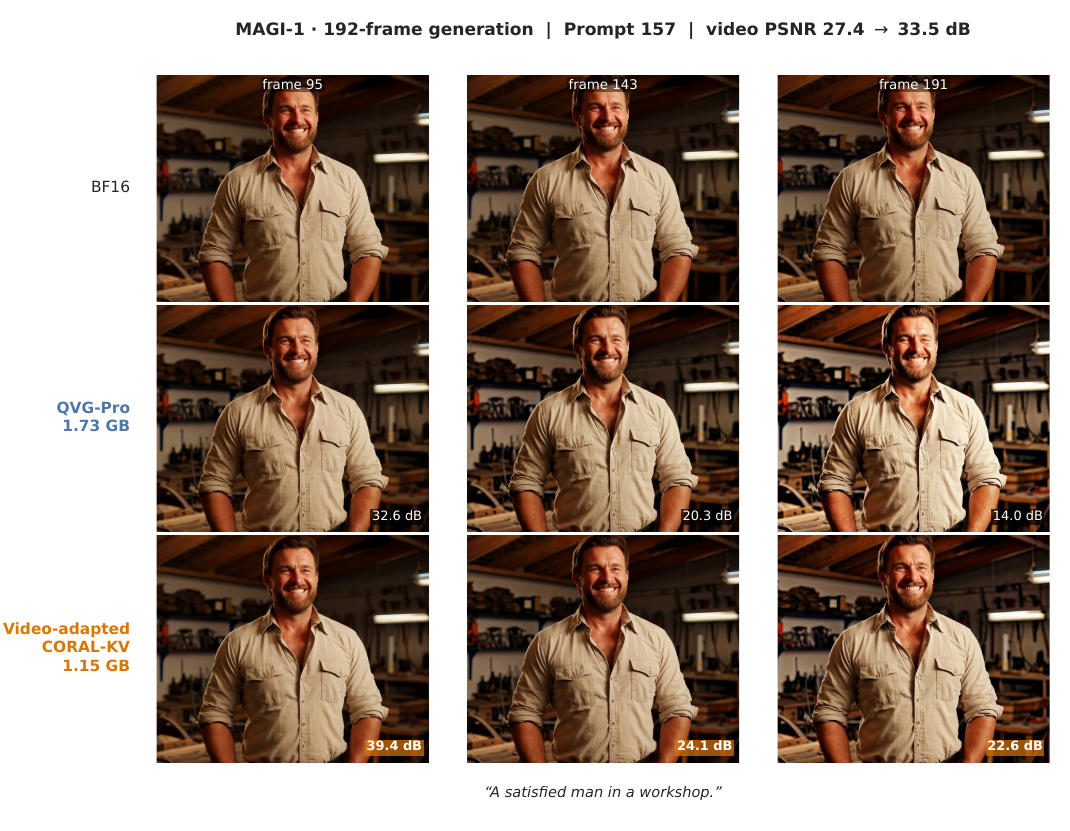}
    \vspace{2pt}
    \includegraphics[width=0.80\textwidth]{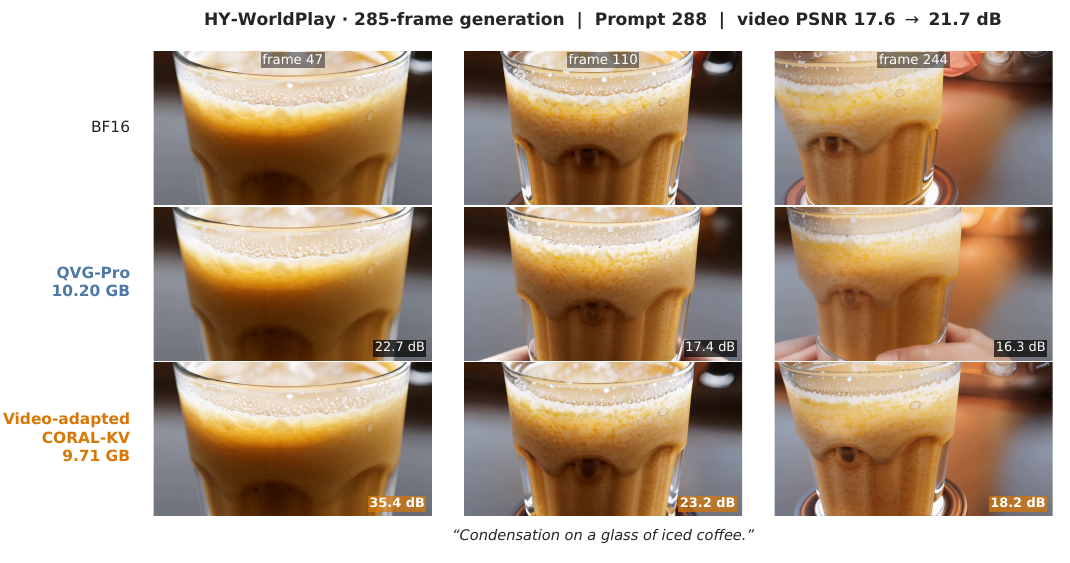}
    \caption{\textbf{Long-range reconstruction examples.}
    MAGI-1 (top; frames 95, 143, and 191) and HY-WorldPlay
    (bottom; frames 47, 110, and 244). Rows compare BF16, QVG-Pro, and
    video-adapted CORAL-KV; tile labels report full-frame PSNR against BF16.}
    \label{fig:additional-long-range}
\end{figure}

\begin{figure}[p]
    \centering
    \includegraphics[width=\textwidth]{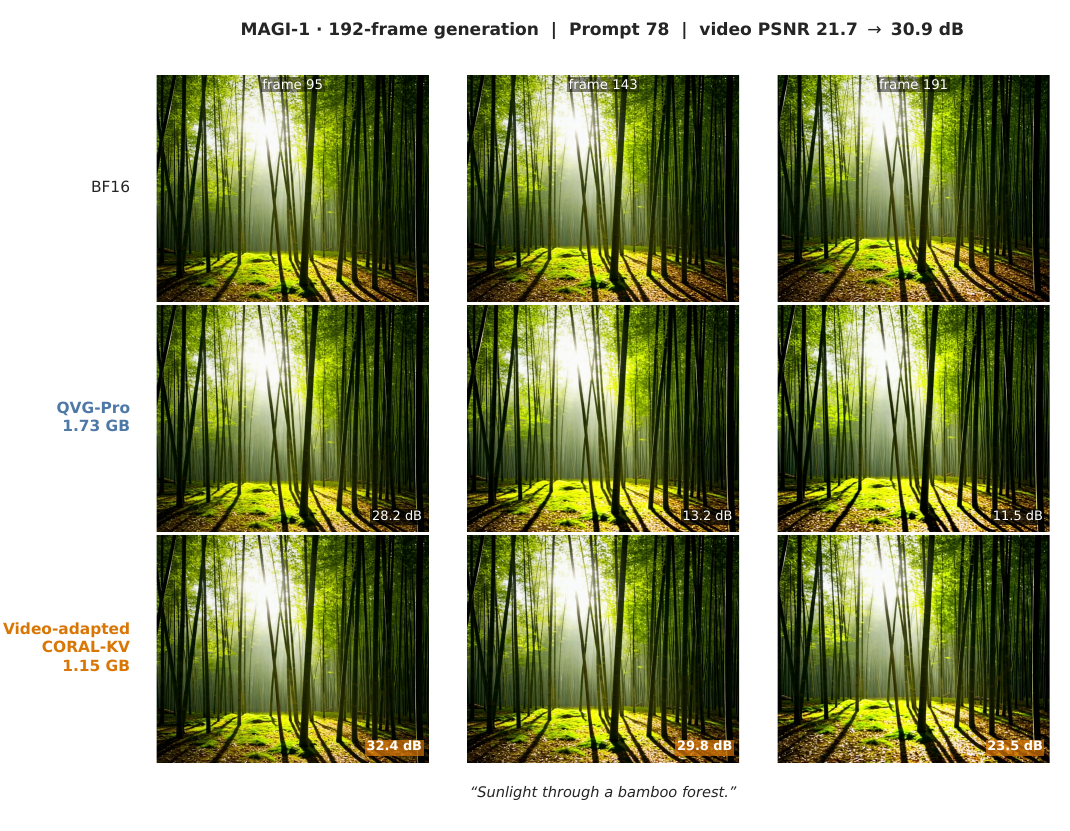}
    \vspace{2pt}
    \includegraphics[width=\textwidth]{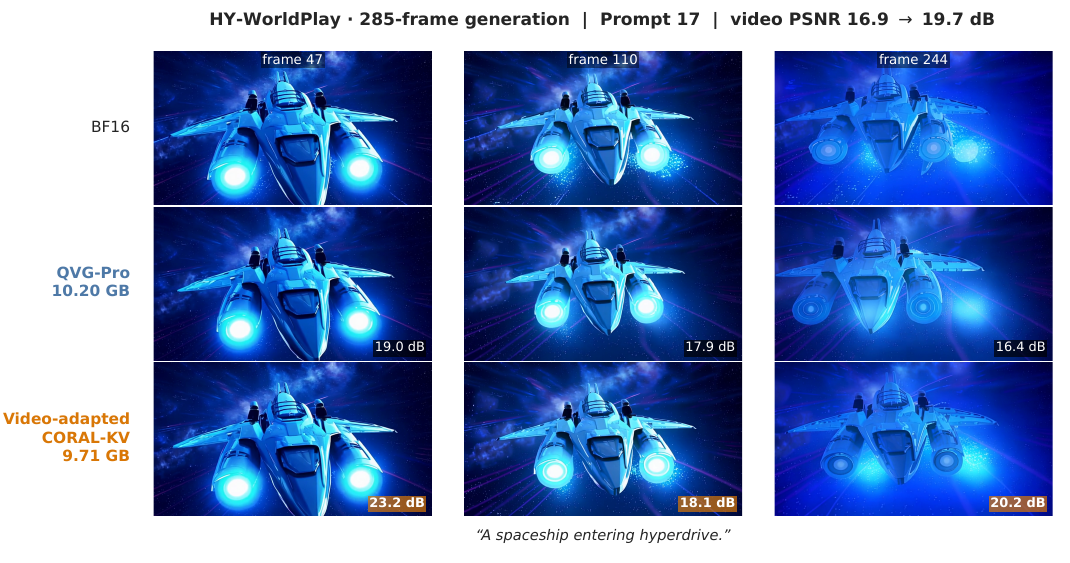}
    \caption{\textbf{Long-range reconstruction examples.}
    MAGI-1 (top; frames 95, 143, and 191) and HY-WorldPlay
    (bottom; frames 47, 110, and 244). Rows compare BF16, QVG-Pro, and
    video-adapted CORAL-KV; tile labels report full-frame PSNR against BF16.}
    \label{fig:additional-long-range-pair-2}
\end{figure}

Figures~\ref{fig:additional-long-range}--\ref{fig:additional-long-range-pair-2}
show illustrative examples of the quality difference between QVG-Pro and
video-adapted \method{}, alongside matched BF16 references.

\FloatBarrier

\begin{figure}[p]
    \centering
    \includegraphics[width=0.78\textwidth]{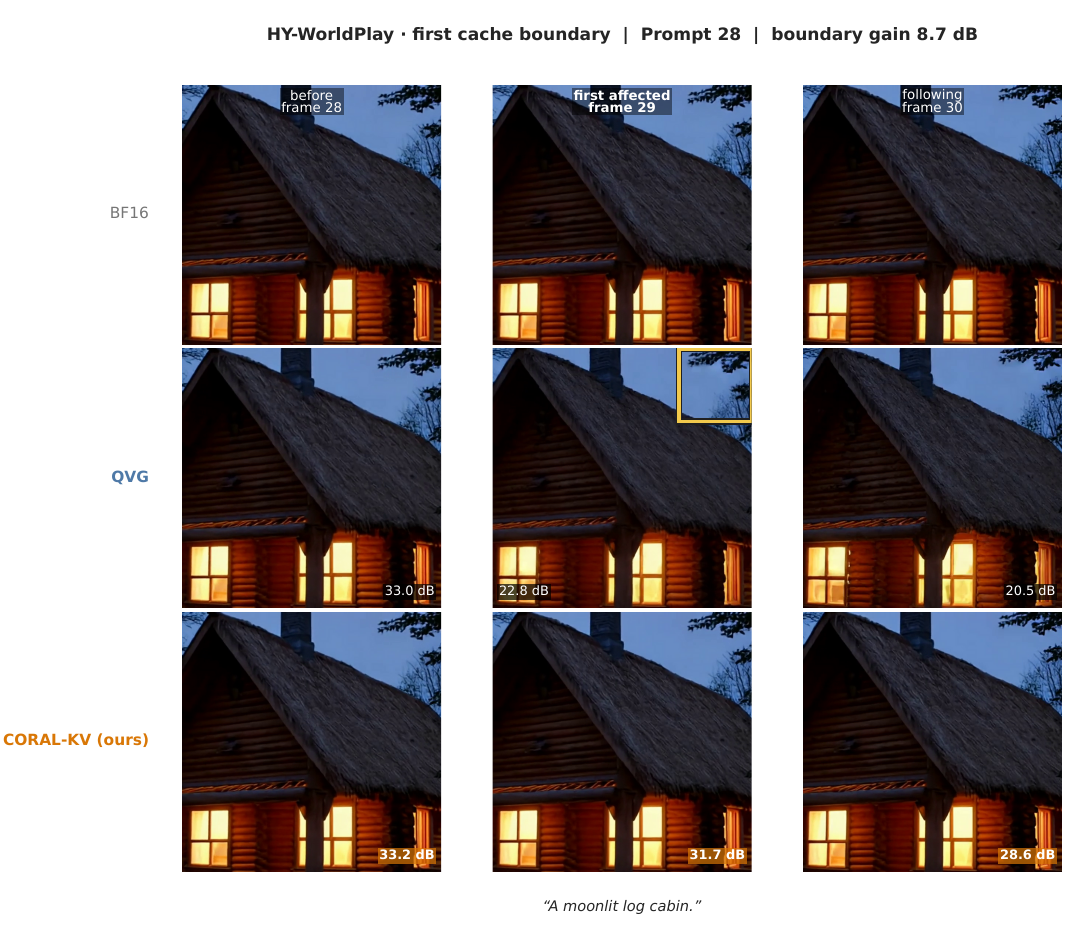}
    \caption{\textbf{HY-WorldPlay first-boundary reconstruction.}
    BF16, QVG, and CORAL-KV at frames 28--30. The yellow box marks QVG's
    largest local fidelity gap at frame 29; tile labels give PSNR versus BF16.}
    \label{fig:additional-boundaries}
\end{figure}

\begin{figure}[p]
    \centering
    \includegraphics[width=0.78\textwidth]{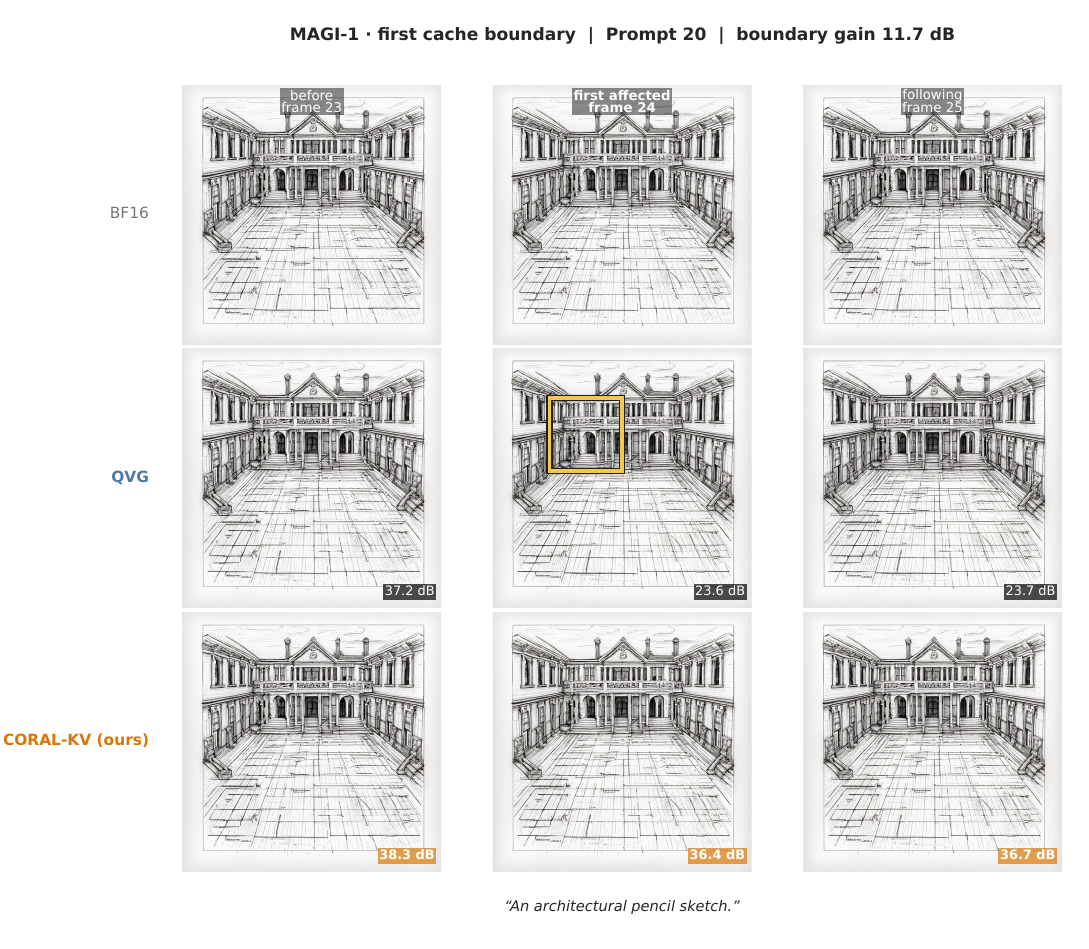}
    \caption{\textbf{MAGI-1 first-boundary reconstruction.}
    BF16, QVG, and CORAL-KV at frames 23--25. The yellow box marks QVG's
    largest local fidelity gap at frame 24; tile labels give PSNR versus BF16.}
    \label{fig:additional-boundaries-magi}
\end{figure}

\end{document}

%% file: math_commands.tex
\usepackage{amsmath,amsfonts,bm}

\def\eqref#1{equation~\ref{#1}}
\def\1{\bm{1}}

\DeclareMathAlphabet{\mathsfit}{\encodingdefault}{\sfdefault}{m}{sl}
\SetMathAlphabet{\mathsfit}{bold}{\encodingdefault}{\sfdefault}{bx}{n}